\documentclass{article} % For LaTeX2e
\usepackage{iclr2022_conference,times}
\usepackage{amsmath}
\usepackage{multirow, multicol}
\newcommand{\pprelu}{\text{PReLU}^+}

\usepackage{graphicx}
\usepackage{subfig}

\usepackage{wrapfig}
\usepackage{amsmath,amsfonts,bm}

\def\eqref#1{equation~\ref{#1}}
\def\1{\bm{1}}

\DeclareMathAlphabet{\mathsfit}{\encodingdefault}{\sfdefault}{m}{sl}
\SetMathAlphabet{\mathsfit}{bold}{\encodingdefault}{\sfdefault}{bx}{n}

\usepackage{hyperref}
\usepackage{url}

\title{Parameterizing Activation Functions for Adversarial Robustness}

\author{Sihui Dai, Saeed Mahloujifar \& Prateek Mittal \\
Princeton University
}

\iclrfinalcopy % Uncomment for camera-ready version, but NOT for submission.
\begin{document}

\maketitle

\begin{abstract}

Deep neural networks are known to be vulnerable to adversarially perturbed inputs.  A commonly used defense is adversarial training, whose performance is influenced by model capacity.  While previous works have studied the impact of varying model width and depth on robustness, the impact of increasing capacity by using \textit{learnable} parametric activation functions (PAFs) has not been studied.
We study how using learnable PAFs can improve robustness in conjunction with adversarial training.
We first ask the question: \textit{how should we incorporate parameters into activation functions to improve robustness?} % Using a set of parametric activation functions (PAFs) which allow us to vary activation function shape from ReLU, we first identify properties of activation function shape which directly influence robustness by measuring adversarial robustness on models trained with standard training.
To address this, we analyze the direct impact of activation shape on robustness through PAFs and observe that activation shapes with positive outputs on negative inputs and with high finite curvature can increase robustness.
%We observe that modifying ReLU to give positive outputs on negative inputs and have high finite curvature increases robustness.
We combine these properties to create a new PAF, which we call Parametric Shifted Sigmoidal Linear Unit (PSSiLU). 
We then combine PAFs (including PReLU, PSoftplus and PSSiLU) with adversarial training and analyze robust performance.  We find that PAFs optimize towards activation shape properties found to directly affect robustness.  Additionally, we find that while introducing only 1-2 learnable parameters into the network, smooth PAFs can significantly increase robustness over ReLU.  For instance, when trained on CIFAR-10 with additional synthetic data, PSSiLU improves robust accuracy by 4.54\% over ReLU on ResNet-18 and 2.69\% over ReLU on WRN-28-10 in the $\ell_{\infty}$ threat model \textit{while adding only 2 additional parameters into the network architecture}.   The PSSiLU WRN-28-10 model achieves 61.96\% AutoAttack accuracy, improving over the state-of-the-art robust accuracy on RobustBench \citep{croce2020robustbench}. %Additionally, we study the learned shape of parametric activation functions. We find that the shapes of parametric activation functions optimize towards the properties found to improve adversarial robustness on standard trained models, suggesting that these properties also improve the model's ability to fit the training data. We also observe that even when the activation function's expressivity is restricted through regularization, parametric activation functions can improve robustness through optimization.
Overall, our work puts into context the importance of activation functions in adversarially trained models.
\end{abstract}

\section{Introduction}
Deep Neural Networks (DNNs) can be fooled by perceptually insignificant perturbations known as adversarial examples ~\citep{szegedy2013intriguing}.  A commonly used approach to defend against adversarial examples is adversarial training \citep{madry2017towards, zhang2019theoretically} which involves training models using adversarial images.  Previous studies have shown that the performance of adversarial training depends on model capacity \citep{madry2017towards}; larger models are able to fit the training set better leading to higher robust accuracy.  These findings raise the question, if adversarial training requires high capacity models, where in the model architecture should we introduce additional parameters? Many studies have observed the impact of factors such as model width and depth \citep{gowal2020uncovering, wu2020does, xie2019intriguing}, but to the best of our knowledge, the potential of increasing capacity through learnable parametric activation functions (PAFs) has not been studied. %\saeed{Just a minor comment: Can we make this last sentence a bit more dramatic? I really like the way we talk about this in the 3rd para of the related work section }
We first ask the question
\begin{quote}
    \textit{How should we parameterize activation functions to improve robustness? }
\end{quote}
%\saeed{Let's start with explaining what the goal is. I think we can say sometihng like: To find the right way of parametrizing the activation function for robustness, we need to first find the most important aspect of an activation function that play a role in robustness....}
To find the right way to parameterize activation functions for robustness, we first need to identify which aspects of activation function shape impact robustness.  We use a set of parametric activation functions (PAFs) with a parameter controlling aspects of shape such as behavior on negative inputs, behavior on positive inputs, and behavior near zero.  We vary the PAF parameter and evaluate the robustness of standard trained models to identify properties of activation function shape that are conducive to robustness. Using standard trained models allows us to decouple the direct impact of activation functions from the impact of adversarial training.
%\saeed{Should we have a sentence summary of why we do standard training instead of adversarial training? I think a our reason could be two fold: 1) computation complexity of adversarial training 2) disentanglement of restrictive bias and preference bias.} 
Surprisingly, we observe a clear trend between activation function shape and robustness for standard trained models.  We find that we can increase robustness by adjusting ReLU to output positive values on negative inputs and to have high finite curvature, the maximum value of the second derivative. 
%\saeed{I think one thing that is missing here is why we suddenly switch to PAFs? In all of our experiments for standard training we used a fixed activation, but we suddenly want to design a PAF. We can say something like this: Although we identify the two properties, these properties are very generic and cannot lead us to a one-fits-all solution for all datasets. For example, we say high finite curvature, but it is not clear how much this curvature should be. Hence, we design a parametric activation function that could be optimized for the best solution that leads to best robustness.}
We combine these properties into a new PAF which we call Parametric Shifted Sigmoidal Linear Unit (PSSiLU) shown in Figure \ref{fig:pssilu}.  PSSiLU uses two parameters $\alpha$ and $\beta$; $\beta$ controls the behavior on negative inputs and $\alpha$ controls curvature.
We then ask the question:
\begin{quote}
    \textit{How do parametric activation functions perform when combined with adversarial training?}
\end{quote}
We train models using PAFs (including PReLU, PSoftplus, and PSSiLU) with learnable parameter with adversarial training and observe the resulting AutoAttack robust accuracy \citep{croce2020reliable} and learned shape of the activation function.  We find that while introducing only 1-2 parameters into the network, certain PAFs can significantly improve robustness over ReLU (Table \ref{tab:intro_sum}).  For instance, when trained on CIFAR-10 with an additional 6M synthetic images from a generative model (DDPM-6M), PSSiLU improves robust accuracy by 2.69\% over ReLU on WideResNet(WRN)-28-10 (and 4.54\% over ReLU on ResNet-18) in the $\ell_{\infty}$ threat model \textit{while adding only 2 additional parameters into the network architecture}.  The WRN-28-10 model achieves 61.96\% robust accuracy, making it the top performing model in its category on RobustBench \citep{croce2020reliable}. We find that PAFs can increase robustness in two ways: 1) through increasing expressivity, allowing the model to better fit the training data, and 2) through optimization, allowing the model to reach a more optimal minimum.  Additionally, we find that activation functions optimize towards the properties observed to increase robustness on standard trained models, suggesting that these properties also allow the model to better fit the data. % Additionally, we observe that even when we restrict the expressivity of the model by regularizing the parameters of the PAF, we can achieve higher robust accuracy through optimization. %Since the performance of adversarial training depends on the optimization landscape of the model, it is important to understand how model architecture impacts the performance of adversarial training.

%One aspect of neural network architecture is activation function.  We ask the questions: \begin{quote}\textit{How do activation functions impact the robustness of adversarially trained models?  What properties of activation functions are most important for robust models?} \end{quote}
%We posit that activation functions can impact adversarially trained models in two ways: (1) through architecture and (2) through adversarial optimization.  By impact through architecture, we refer to the direct impact of activation function shape on robustness without the use of adversarial training.  This includes how factors such as curvature impact the robustness of the model.  By impact through adversarial optimization, we refer to the impact of the optimization landscape induced by the activation function and the question of whether this landscape allows for the model to converge to a more optimal minimum.  In our evaluations, we work with \textit{parametric activation functions} (PAFs) with a parameter that controls shape.  This allows us to experiment across a wide variety of activation function shapes which have not been studied in prior works.

 \begin{minipage}{\textwidth}
  \begin{minipage}[b]{0.63\textwidth}
    \centering
    
    \vspace{-20pt}
    \includegraphics[width=\textwidth]{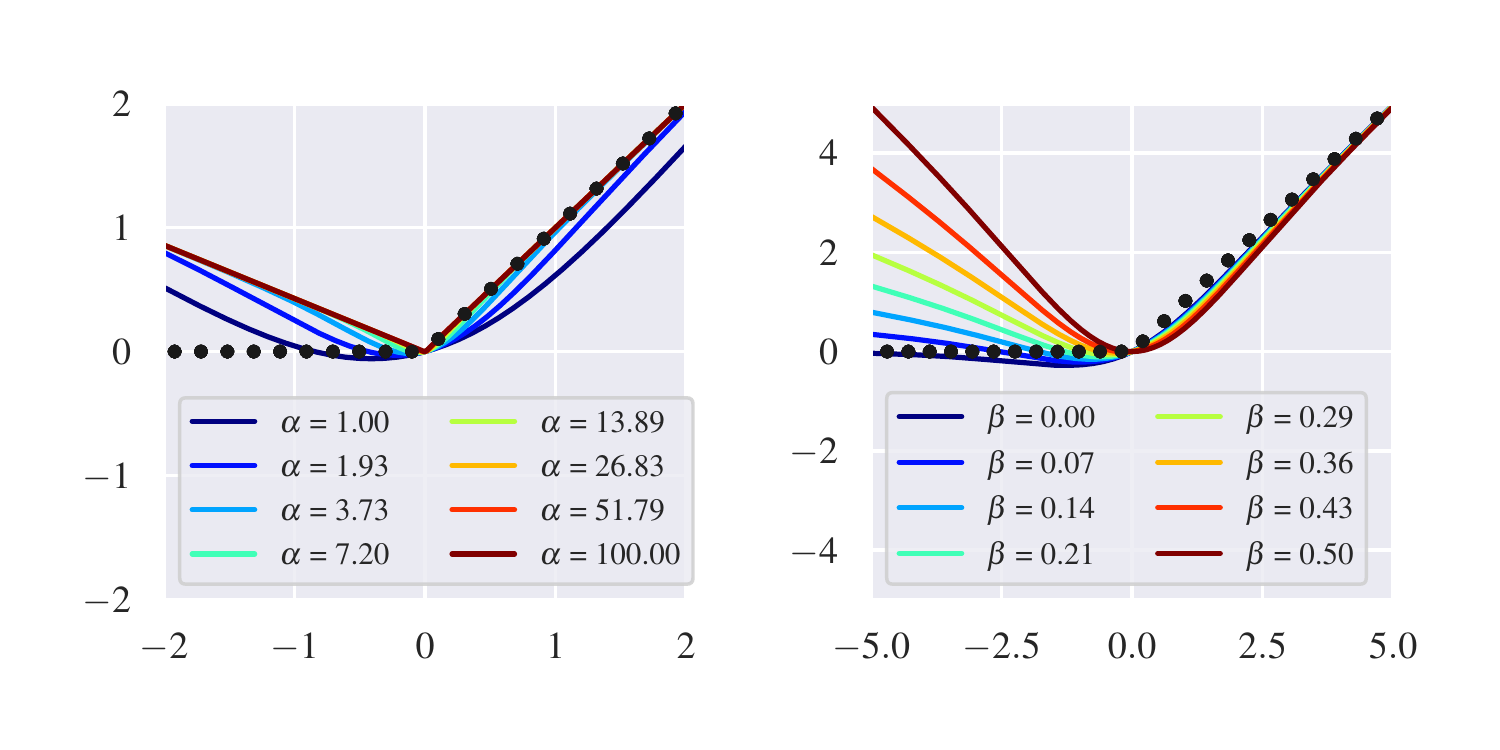}
    \label{fig:pssilu}
    \vspace{-25pt}
    \captionof{figure}{Shape of PSSiLU at various values of $\alpha$ and $\beta$.  Left: $\beta$ is fixed to 0.3 while $\alpha$ is varied. Right: $\alpha$ is fixed to 1 while $\beta$ is varied.  ReLU is given by the dotted black line.  We can see that $\alpha$ controls the curvature of the function near 0 while $\beta$ controls the behavior on negative inputs.}
  \end{minipage}
  \hfill
  \begin{minipage}[b]{0.35\textwidth}
    \centering
    \begin{tabular}{|c|cc|}\hline
      & Activation & AA \\ \hline
        \multirow{3}{*}{\rotatebox[origin=c]{90}{\parbox[c]{1cm}{\centering Non}}} 
         &Softplus & 49.14 \\
        &SiLU & 55.10 \\
        &ReLU & 59.27 \\
        \hline
        \multirow{3}{*}{\rotatebox[origin=c]{90}{\parbox[c]{1cm}{\centering PAF}}} &PSoftplus & 60.94 \\
        &PSiLU & 60.37 \\
        &PSSiLU & 61.96 \\
        \hline
      \end{tabular}
      \captionof{table}{AutoAttack robust accuracy of WRN-28-10 models trained using PGD adversarial training on CIFAR-10+DDPM-6M.  Nonparametric activation functions are labeled by ``Non".  Parametric activation functions are labeled by ``PAF".}
      \label{tab:intro_sum}
    \end{minipage}
  \end{minipage}

In summary, our contributions are as follows
\begin{enumerate}
    \item  We find that in the absence of adversarial training, certain properties of activation function shape (namely positive outputs on negative inputs and high finite curvature) are correlated with robustness of standard trained models against a weak adversary.  This suggests that these properties of shape have a direct influence on robustness.  %We investigate which properties in activation function shape are conducive to robustness by evaluating standard trained models using PAFs of fixed parameter. We find that robustness increases when ReLU is modified to give positive output on negative inputs and have high finite curvature. %Using PAFs with fixed parameter, we evaluate standard trained models against a weak adversary at various shapes of activation function.  We find that robustness increases when ReLU is modified to give positive output on negative inputs and have high finite curvature. 
    Using these observations, we introduce a new PAF which we call PSSiLU with two parameters which control these properties (Figure \ref{fig:pssilu}).
    \item While prior works only explore the use of activation functions with fixed parameter with adversarial training, we explore the use of PAFs with \textit{learnable} parameter and observe their impact on robustness with adversarial training.  We find that PAFs optimize towards the same properties which improve robustness in the standard training setting, suggesting that these properties also allow the model to better fit the training data.
    \item We unlock the full potential of using learnable PAFs with adversarial training by training with additional synthetic data, increasing robust accuracy over nonparametric activation functions (Table \ref{tab:intro_sum}).  The PAFs tested only add 1-2 parameters into the entire network (all parameters of PAFs are shared across all activations), but we find that smooth PAFs are able to improve robust accuracy over ReLU and other nonparametric activation functions.  This emphasizes the importance of considering activation functions in adversarial training.
    %We combine PAFs with adversarial training and observe the robust accuracy and learned shapes of PAFs.  We find that PAFS optimize towards the same properties which improve robustness in the standard training setting, suggesting that these properties also allow the model to better fit the training data.  We unlock the potential of using PAFs with adversarial training by training with additional synthetic data, increasing robust accuracy over nonparametric activation functions (Table \ref{tab:intro_sum}).  We find that even without additional data, smooth PAFs can outperform ReLU.
    \item We find that when trained on CIFAR-10 with additional synthetic data, PSSiLU improves by 2.69\% over ReLU on WRN-28-10 (and 4.54\% over ReLU on ResNet-18) in the $\ell_{\infty}$ threat model, making it the top performing model in its category on RobustBench \citep{croce2020robustbench}.  Additionally, we find that the family of activation functions captured by PSSiLU consistently achieves high robust accuracy, outperforming RELU across multiple datasets, architectures, perturbation types, and sources of additional data.
    %\item We investigate the shape of the PSSiLU model and find that the regularized parameter $\beta$ learns to take value 0, which makes PSSiLU equivalent to Parametric SiLU (PSiLU), another PAF studied in this paper.  We find that despite learning to be PSiLU, PSSiLU improves over the performance of PSiLU by 3.01\% on ResNet-18 and 1.59\% on WRN-28-10 in the $\ell_{\infty}$ threat model, suggesting that parametric activations can also improve the performance of adversarial training by improving optimization, allowing the model to find a more optimal local minimum.
\end{enumerate}

\vspace{-5pt}
\section{Related Works}
\textbf{Adversarial Attacks and Adversarial Training.}
Previous studies have shown that modern NNs can be fooled by perturbations known as adversarial attacks, which are imperceptible to humans, but cause NNs to predict incorrectly with high confidence \citep{szegedy2013intriguing}.  These attacks can be generated in a white box \citep{goodfellow2014explaining, madry2017towards, carlini2017towards, croce2020reliable} or black-box \citep{brendel2017decision, andriushchenko2020square, papernot2016transferability} manner. 

Adversarial training is a defense in which adversarial images are used to train the model. The first variant of adversarial training is PGD adversarial training \citep{madry2017towards}.  Since then other variants of adversarial training have been introduced to improve robust performance \citep{wang2019dynamic, zhang2020geometry} and reduce tradeoff between natural and robust accuracy\citep{zhang2019theoretically, zhang2020fat, wu2020adversarial}. Recent works have also explored how to improve robustness when combined with adversarial training \citep{gowal2020uncovering, pang2020bag}.  These include techniques such as using additional data \citep{carmon2019unlabeled, sehwag2021improving, rebuffi2021fixing}, and early stopping \citep{rice2020overfitting}. \citet{croce2020robustbench} provide a leaderboard for ranking defenses against adversarial attacks, and currently the top defenses on this leaderboard are all based on adversarial training.

\textbf{Importance of Model Capacity in Adversarial Training.} Prior works have indicated that the performance of adversarial training depends on model capacity.  \citet{madry2017towards} demonstrated that large model capacity is necessary for adversarial training to successfully fit the training data.  Recently, \citet{bubeck2021universal} proved that $nd$ parameters are necessary for a model to robustly fit $n$ $d$-dimensional data points.  These findings raise the question, if adversarial training requires high capacity models, where in the model architecture should we introduce additional parameters?  In line with this question, multiple works have studied the impact of changing the capacity of DNNs by modifying width and depth on robustness \citep{wu2020does, xie2019intriguing, gowal2020uncovering}.  However, the question of how introducing parameters into activation functions impacts robustness has been unexplored.  We address this question by observing the performance of parametric activation functions in conjunction with adversarial training.

\textbf{Activation Functions and Robustness.} While most works on activation functions focus on improving natural accuracy \citep{clevert2015fast, glorot2011deep, ramachandran2017searching, he2015delving}, there have been a few works which explore activation functions in the adversarial setting.  One line of works evaluates the impact of properties such as boundedness \citep{zantedeschi2017efficient}, symmetry \citep{zhao2016suppressing}, data dependency \citep{wang2018adversarial}, learnable shape \citep{tavakoli2020splash}, and quantization \citep{rakin2018defend} on robustness without using adversarial training.  A more closely related line of works evaluates the performance of models using various nonparametric activation functions in conjunction with adversarial training \citep{xie2020smooth, gowal2020uncovering, singla2021low}. % When combined with adversarial training, activation functions can impact the model through the optimization landscape of adversarial training. 

In contrast to prior works, we experiment with \textit{parametric activation functions} (PAFs), allowing us to explore a wider range of activation function shapes and understand the impact of increasing model capacity through activation functions.  We will first identify qualities of activation functions which have a direct impact on robustness by observing standard trained models in order to design a PAF to use with adversarial training (Section \ref{sec:directeffect}).  We then combine PAFs with \textit{learnable parameter} with adversarial training and analyze their potential in improving robust accuracy obtained through adversarial training (Section \ref{sec:adv_train}).

\section{Searching for a good parameterization}
\label{sec:directeffect} 
Existing PAFs are designed for improving natural accuracy through standard training without considering robustness, leading to the question: \textit{how should we design a PAF for improving robustness?}  One challenge in designing PAFs for robustness is that there are many shapes that an activation function can take, leading to a large design space.  Since ReLU is commonly used in DNNs and has good performance, we choose a set of 6 different PAFs which can take on the shape of ReLU while allowing us to vary behavior from ReLU, which we discuss in Section \ref{sec:activations}.  By controlling the shape of these PAFs, we identify qualities of activation functions that are conducive to robustness.

This raises another question: \textit{How should we measure what parameterizations are considered good?} 
We can divide the impact of activation functions on robustness into the direct impact of activation function shape (restriction bias) and the impact of activation functions in conjunction with optimization through adversarial training (preference bias).  In this section, we focus on the restriction bias by measuring the robustness of standard trained models against a weak adversary in Section \ref{sec:identifyingparam}.  We use a weak adversary since standard trained models are not robust against strong attacks.  By observing standard trained models, we identify properties in activation function shape that directly affect robustness.  We then combine the observed properties into a novel PAF in Section \ref{sec:pssilu} for use with adversarial training (Section \ref{sec:adv_train}).

\begin{figure}[t]
    \centering
    \includegraphics[width=0.85\textwidth]{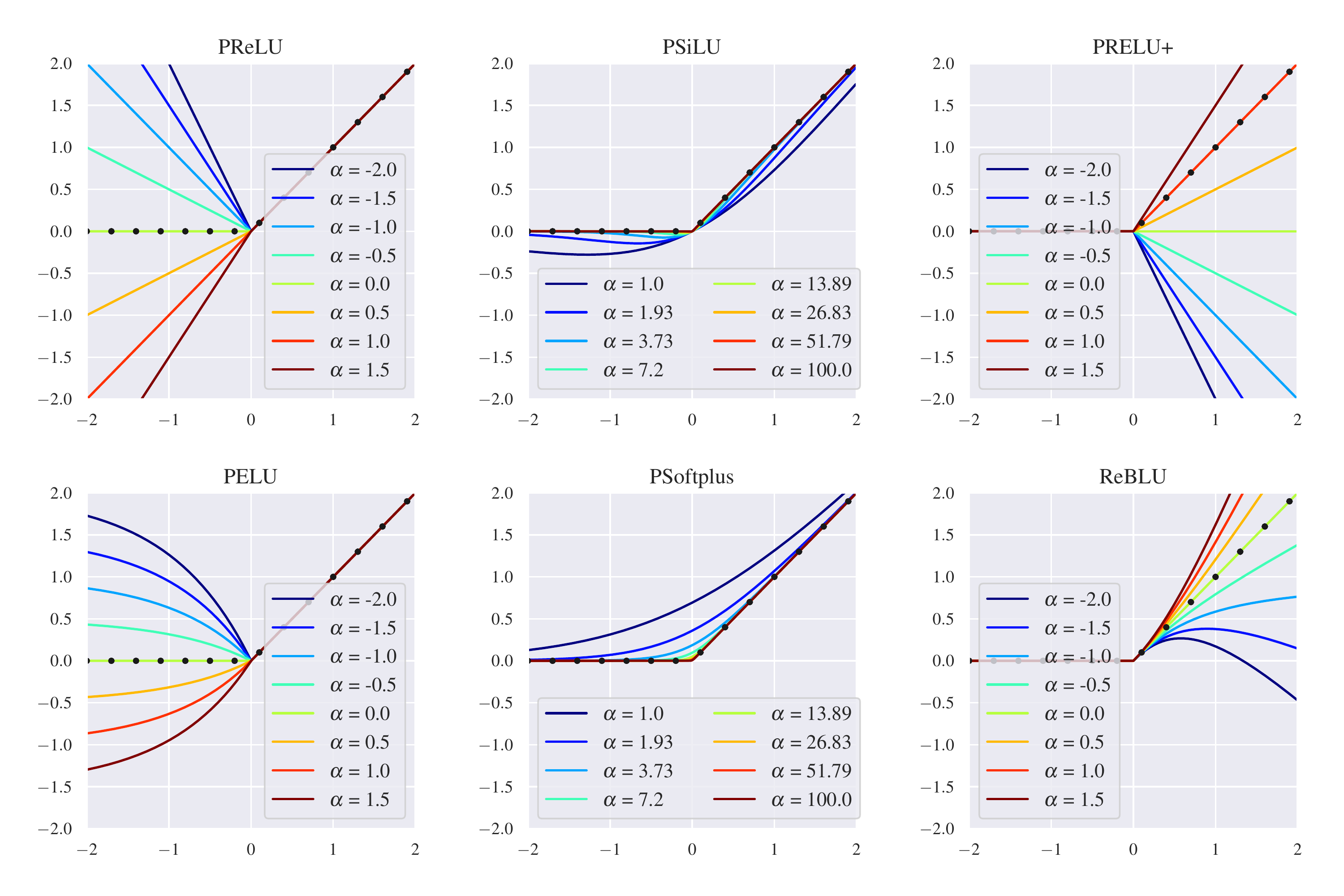}
    \caption{Visualization of parametric activation functions at various values of parameter $\alpha$.}
    \label{fig:act_plots_all}
\end{figure}

\subsection{Activation Function Search Space}
\label{sec:act_def}
\label{sec:activations}
Since ReLU is commonly used in DNN architectures, we first consider a set of PAFs with a single parameter $\alpha$ that are able to model the shape of ReLU, while also allowing for variation in behavior at different regimes in the input. We divide our initial set of PAFs into 3 groups: those which capture variation on negative inputs, those which capture variation for inputs of small magnitude, and those which capture variation for large positive inputs.  The shapes at varied values of parameter $\alpha$ of all activation functions that will be introduced are shown in Figure \ref{fig:act_plots_all}.

To capture variation on negative inputs, we consider parametric ReLU (PReLU) \citep{he2015delving} and parametric ELU (PELU) \citep{clevert2015fast} defined as follows:
  \begin{align}
    \text{PReLU}_{\alpha}(x) = \begin{cases} 
      \alpha x & x \le 0 \\
      x  & x > 0
   \end{cases} &&
   \text{PELU}_{\alpha}(x) = \begin{cases} 
      \alpha(e^x - 1) & x \le 0 \\
      x  & x > 0
   \end{cases} 
  \end{align}

To capture variation for inputs near zero, we consider two continuous parametric activation functions parametric SiLU (PSiLU) \citep{ramachandran2017searching} and parametric Softplus (PSoftplus) \citep{dugas2001incorporating}.  These activation functions are defined as follows:
\begin{align}
    \text{PSiLU}_{\alpha}(x) = x \sigma(\alpha x) &&
    \text{PSoftplus}_{\alpha} = \frac{1}{\alpha}\log(1 + e^{\alpha x})
\end{align}
where $\sigma(x) = \frac{1}{1 + e^{-x}}$ is the sigmoid function.

To capture variation on positive inputs, we introduce two activation functions: one which we call Positive PReLU ($\pprelu$) and the other which we call Rectified BLU (ReBLU).  $\pprelu$ has a parameter controlling the slope of the linear portion of ReLU.  ReBLU allows for nonlinear behavior on positive inputs and is based off Bendable Linear Unit (BLU)  defined as $\text{BLU}_{\alpha} = \alpha (\sqrt{x^2 + 1} -1 ) + x$ \citep{8913972}.  To allow BLU to take the shape of ReLU for comparison, we modify BLU so that it is piecewise and outputs 0 for all negative inputs. We define $\pprelu$ and ReBLU as follows:
\begin{align}
    \pprelu_{\alpha}(x) = \begin{cases} 
       0 & x \le 0 \\
      \alpha x  & x > 0
   \end{cases}
&&
   \text{ReBLU}_{\alpha}(x) = \begin{cases} 
      0 & x \le 0 \\
      \text{BLU}_{\alpha}(x)  & x > 0
   \end{cases} 
\end{align}
\label{PPReLU_PBLU_def}
\vspace{-5pt}
\subsection{Identifying parameters conducive to robustness on standard trained models}
\label{sec:identifyingparam}
Using our set of 6 PAFs, we vary the shapes of activation functions and measure their impact on robustness in order to design a new PAF which models more robust activation function shapes.  To disentangle the impact of activation function shape from the impact of training, we analyze the robustness of standard trained models. %\saeed{It's not clear if the focus of this sentence is "standard training" or "weak adversary".  I thin we should first say we do standard training to disentangle. Then explain we use weak attack because the with strongest attacks its hard to see the robustness benefits on standard models.}.  
We use a weak adversary in this analysis because standard trained models are not robust; using a strong adversary would make it difficult to see the impact of activation function shape.  Note that we do not expect the standard trained model to achieve any robustness against strong attacks even with PAFs. However, in Section \ref{sec:adv_train}, we will experiment with adversarially trained models against a strong adversary. %\saeed{I think this sentence is crucial to avoid confusion. We can expand it and say: Note that we do not expect the normally trained model to get any robustness against strong attacks, even enhanced activation functions. However, in section ...}.

\begin{figure}[t]
    \centering
    \includegraphics[width=\textwidth]{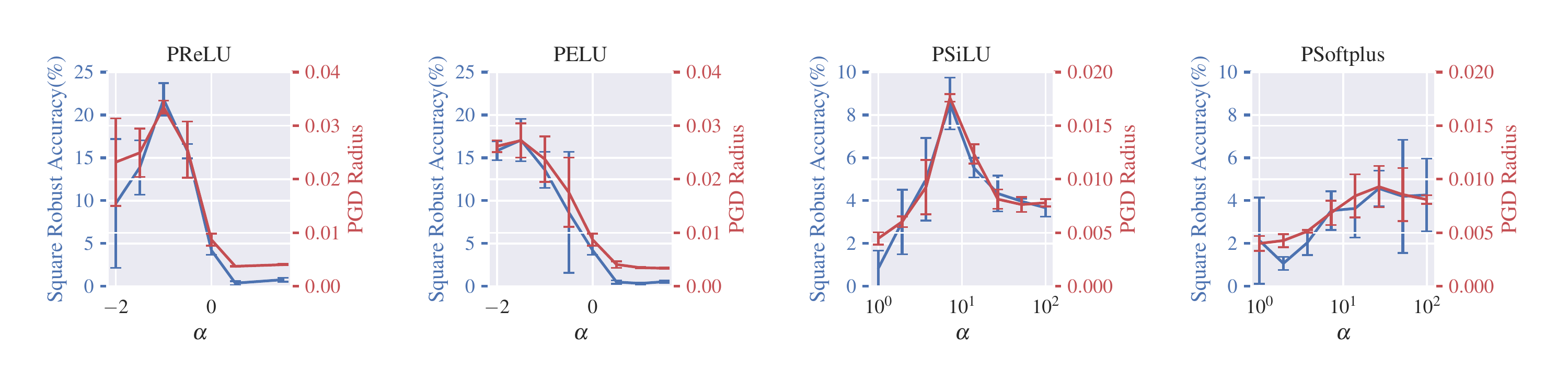}
    \vspace{-20pt}
    \caption{Square robust accuracy and average minimum PGD radius for ResNet-18 models trained on CIFAR-10 with various parameter $\alpha$.  Results are computed over 3 trials. Red points indicate the measured PGD radius while blue points indicate the Square robust accuracy across models.}
    \label{fig:resnet_18_std}
\end{figure}

%\saeed{This para could benefit from a heading.}
For our weak adversary, we use both white box and black box attacks.  For white box attack, we search for the smallest perturbation radius that leads to misclassification on 4-step PGD perturbed inputs. For black box attacks, we use a query-restricted black-box adversary (Square with 1000 queries \citep{andriushchenko2020square, croce2020reliable}) and measure the robust accuracy of the models on adversarial examples. %These attacks give us a quantitative measurement for robustness even on standard models and allow us to better compare the impact of shape of activation functions.
%\saeed{We can specify the common property of these two attacks: they both give us a quantitative measurement for robustness on a given point x, whereas the strongest attacks will only give a 0-1 value because a point is either robust or non-robust. This will help us in having a better comparison.}  
Additional details on experimental setup are located in Appendix \ref{app:standardsetup}.  We present the measured Square robust accuracy and PGD radius for ResNet-18 models trained on CIFAR-10 at various parameter $\alpha$ in Figure \ref{fig:resnet_18_std}.

From Figure \ref{fig:resnet_18_std}, we observe clear trends across PReLU, PELU, PSiLU, and PSoftplus and find that Square robust accuracy and PGD radius are highly correlated, suggesting that these trends are not the result of obfuscated gradients \citep{athalye2018obfuscated}.  Additionally, we found that these trends generalize to other model architectures and datasets.  We provide results for ResNet-18 models on CIFAR-100 and ImageNette datasets and results for WRN-28-10 and VGG-16 models trained on CIFAR-10 in Appendix \ref{app:generalization}.  We did not observe clear, consistent trends for $\pprelu$ and ReBLU, suggesting that the behavior of activation functions on positive inputs is less important to robustness, but provide the results for these activation functions in Appendix \ref{app:posbehavior}. %\saeed{Instead of saying this, should we dedicate a paragraph that says that the behavior of the activation on positive points is not as important as the curvature and behavior on negative inputs? Therefore, we don't have a dedicatetd parameter to parametarize the behavior of positive point in our design? This makes sense because we say the point of this study was to find the important traits of activations...}

\textbf{Positive outputs on negative inputs increases robustness to weak adversary.}  From Figure \ref{fig:resnet_18_std}, we observe similar trends across both PReLU and PELU.  For these activation functions, $\alpha$ controls the behavior on negative inputs.  We observe a significant increase in robustness for models at $\alpha < 0$ compared to $\alpha > 0$, although there a decrease in robustness when $|\alpha|$ becomes large.  For both PReLU and PELU, $\alpha = 0$ corresponds to ReLU, $\alpha > 0$ outputs negative values for negative inputs, and $\alpha < 0$ outputs positive values on negative inputs (See Figure \ref{fig:act_plots_all}).  This trend suggests that designing a PAF with a parameter that controls the shape of the activation function on negative inputs may be conducive for robustness.  Specifically, we will add a parameter which allows for the PAF to vary the magnitude of positive outputs on negative inputs in a way that is similar to PReLU or PELU (Section \ref{sec:pssilu}).

\textbf{Higher curvature increases robustness to weak adversary.}  From Figure \ref{fig:resnet_18_std}, we observe similar trends between PSiLU and PSoftplus.  For these activation functions $\alpha$ controls the curvature, the maximum value of the second derivative. As $\alpha$ increases, the curvature also increases.  First, we note that the models with highest robustness have $\alpha > 1$, where $\alpha = 1$ models their nonparametric variants commonly used in training neural networks.  At higher values of $\alpha$, the shapes of these activation functions grow close to the shape of ReLU which has curvature of infinity.  We find that for both PSiLU and PSoftplus, robustness initially increases as $\alpha$ increases and then decreases after a certain point, with this trend being highly significant for PSiLU.  This suggests that designing a PAF with a parameter that controls the curvature of the activation function similar to PSiLU and PSoftplus may also benefit robustness.

\subsection{Putting it together: PSSiLU} 
\label{sec:pssilu} \begin{wrapfigure}{r}{0.5\textwidth}
    \vspace{-30pt}
    \centering
    \includegraphics[width=0.5\textwidth]{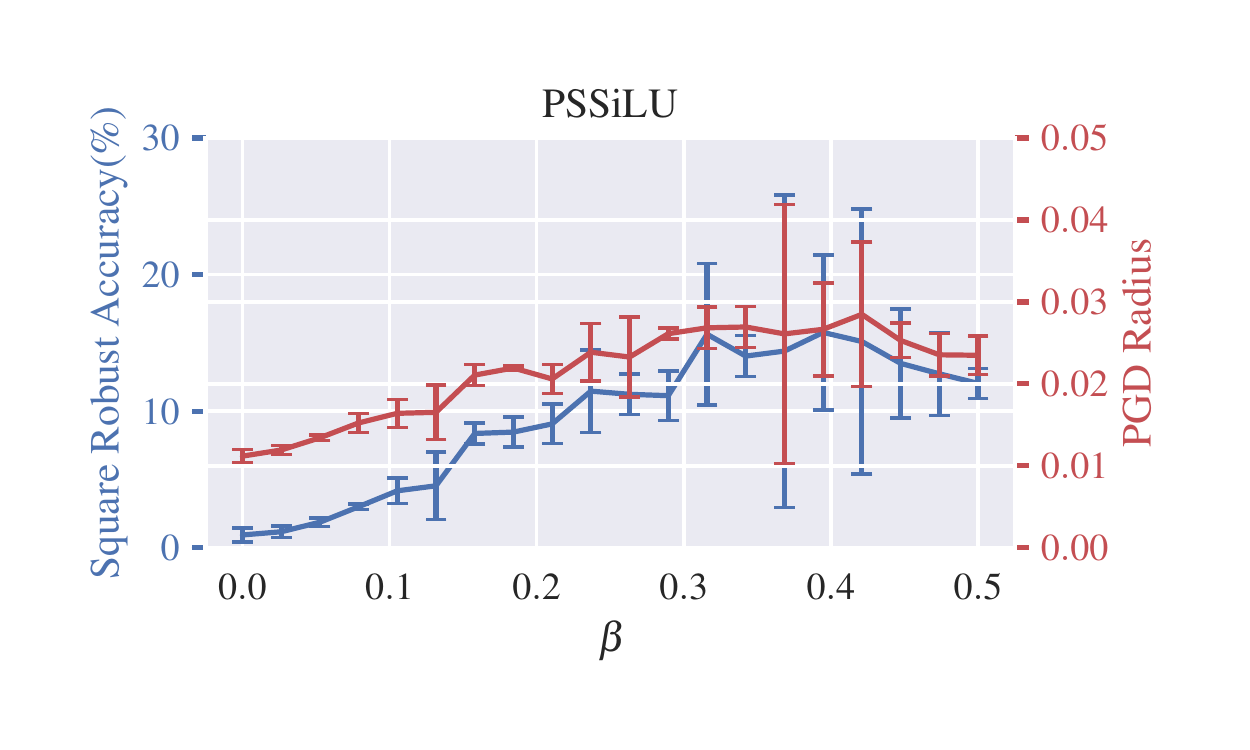}
        \vspace{-30pt}

    \caption{Square robust accuracy and average minimum PGD radius for SSiLU at varied $\beta$ for ResNet-18 models trained on CIFAR-10.  Results are computed over 3 trials.}
    %\vspace{-5pt}

   \label{fig:ssilu_beta}

\end{wrapfigure}
We now combine the properties observed to enhance robustness into an activation function.  One PAF exhibiting both properties is PSoftplus; however the trend in robustness was not as significant for PSoftplus as for PSiLU, PReLU, or PELU.  Since we observed a more significant trend in robustness over parameter $\alpha$ for PSiLU, we introduce a new activation function that is based off of PSiLU.  We call this new activation function Parametric Shifted SiLU (PSSiLU) defined as:
\vspace{-10pt}

\begin{equation}
    \text{PSSiLU}_{\alpha, \beta}(x) = x (\sigma(\alpha x) - \beta) / (1 - \beta)
\end{equation}

where $\alpha, \beta > 0$, $\beta < 1$, and $\sigma(x) = \frac{1}{1 + e^{-x}}$ is the sigmoid function.  At $\beta = 0$, PSSiLU's behavior matches that of PSiLU.  Recall that the shape of PSSiLU at various values of $\alpha$ and $\beta$ were shown in Figure \ref{fig:pssilu}.  $\alpha$ controls curvature around 0 while $\beta$ controls behavior on negative inputs.  Increasing $\beta$ allows PSSiLU's output on input $x < 0$ to grow with the magnitude of $x$ similar to PReLU. 

We perform a similar evaluation of robustness across the $\beta$ parameter of PSSiLU with $\alpha=1$ in Figure \ref{fig:ssilu_beta}.  We find that as $\beta$ increases, there is an increase in robustness, which plateaus at around $\beta = 0.3$.  Since higher values of $\beta$ correspond to giving positive output on negative inputs, this trend matches that of PReLU and PELU. In Appendix \ref{app:pssilu_gen}, we demonstrate that this trend is consistent across architectures (WRN-28-10 and VGG-16) and datasets (CIFAR-100). %\sdcomment{todo in appendix: add additional results for other datasets and architectures}

%In Appendix \ref{app:searchbeta}, we search for the $\beta$ that maximizes robustness on SSiLU and find that it occurs at about $\beta = 0.3$.  For the rest of this paper, we will fix $\beta$ to this value. 

\section{Investigating the performance of adversarially trained models using parametric activation functions}
\label{sec:adv_train}
%\sdcomment{expand to big picture, draw more interesting research performance}
We now combine PAFs with adversarial training to investigate the impact of incorporating parameters into activation functions on adversarial training.  We experiment with the activation functions from Section \ref{sec:activations} with learnable parameters.  Specifically, we add $\alpha$ (and $\beta$ for PSSiLU) to the parameter set $\theta$ that we optimize during adversarial training. We share PAF parameters across all layers in the network, so that PSSiLU only introduces two additional parameters into the model while all other PAFs introduce one new parameter.  We also train models using the commonly used nonparametric activation functions: ReLU, ELU, SiLU, Softplus.  ELU, SiLU, and Softplus correspond to $\alpha=1$ for PELU, PSiLU, and PSoftplus respectively.

We perform experiments on WRN-28-10, ResNet-18, VGG-16 architectures and on CIFAR-10, CIFAR-100, and Imagenette datasets.  For CIFAR-10, we also experiment with using additional data during training.  For additional CIFAR-10 data sources, we use DDPM-6M, a set of 6M CIFAR-10 images generated by DDPM, a generative model \citep{ho2020denoising} and TinyImages-500K (TI-500K), a subset of 500K images from TinyImages \citep{carmon2019unlabeled}.  We note that DDPM-6M does not require any additional real data since DDPM is trained directly on CIFAR-10. DDPM-6M and TI-500K have been shown to improve the robustness of adversarially trained models \citep{carmon2019unlabeled, sehwag2021improving, rebuffi2021fixing}. For the bulk of our experiments, we use 10-step PGD adversarial training \citep{madry2017towards} and focus on $\ell_{\infty}$ attacks, but we include results with TRADES adversarial training \citep{zhang2019theoretically} and $\ell_2$ attacks in the Appendix.  We include additional details about experimental setup in Appendix \ref{app:adversarialsetup}.

\subsection{The importance of additional data}
%\sdcomment{begin with analyzing results on CIFAR-10 first}
We present results for $\ell_{\infty}$ attacks on ResNet-18 in Table \ref{tab:resnet-18_adv_summary} and $\ell_{\infty}$ attacks on WRN-28-10 in Table \ref{tab:wrn_adv_summary}.  We also report results for VGG-16, ResNet-18 models trained with TRADES \citep{zhang2019theoretically}, and ResNet-18 models under an $\ell_2$ adversary in Appendix \ref{app:additional_adv}.

We find that when trained on CIFAR-10 without extra data, most PAFs are unable to outperform their nonparametric variant.  For example, on ResNet-18 (Table \ref{tab:resnet-18_adv_summary}), PELU has 3.20\% lower robust accuracy than ELU when trained on CIFAR-10.  Similarily, PSiLU obtains 1.9\% lower robust accuracy compared to SiLU on CIFAR-10.  This trend can also be seen for WRN-28-10 (Table \ref{tab:wrn_adv_summary}).  Since parametric activation functions are able to take the shape of their nonparametric variants, this suggests that PAFs may need additional regularization to allow the model to converge to a better minimum.

We find that additional data, which can be synthetic as in the case of DDPM-6M, helps regularize PAFs during training, allowing PAFs to outperform their nonparametric variants. For instance, we find that when trained with additional DDPM-6M data, PELU outperforms ELU by 1.70\% and PSiLU outperforms SiLU by 1.13\% on ResNet-18.  A similar trend also holds for WRN-28-10, where PELU outperforms ELU by 8.11\% and PSiLU outperforms SiLU by 5.27\%. %Additionally, we find that with additional data, smooth PAFs can consistently outperform ReLU.  For example, PSoftplus outperforms ReLU by 3.11\% and PSiLU outperforms ReLU by 1.53\%.
\begin{table}[ht]
\vspace{-10pt}

\centering
\resizebox{\textwidth}{!}{
\subfloat[ResNet-18]{
\label{tab:resnet-18_adv_summary}
\begin{tabular}{ |c|c|c|c|c| } 
\hline
& \multicolumn{2}{|c|}{\textbf{CIFAR-10}} &  \multicolumn{2}{|c|}{\textbf{+DDPM-6M}} \\
\hline
Activation & Natural & AA & Natural & AA \\
\hline
  ReLU & 82.29 & 44.58 &  82.83 & 53.67\\
  PReLU & 80.16  & 43.53 & 83.27 & 53.66\\
    \hline

  ELU & 81.85 & \textcolor{violet}{46.76}& 82.47  & 51.59\\
  PELU  & 80.37  &43.56 & 83.07 & 53.29 \\
  \hline
  Softplus &80.46 &\textcolor{violet}{44.64} & 79.44 &49.41 \\
  PSoftplus & 80.40 & 44.48 &84.56 &\textcolor{violet}{56.78} \\
  \hline
  $\pprelu$ & 79.77 & 42.34 & 83.63 & \textcolor{violet}{54.21}\\
  ReBLU & 81.19 & \textcolor{violet}{44.91} & 83.64  & \textcolor{violet}{53.74} \\
  \hline
%  SSiLU & 78.9 & 44.0 & 84.2 & 55.7 & 86.7 & 60.2\\
  SiLU & \textbf{82.53 } & \textbf{\textcolor{violet}{46.78}} & 83.53  & \textcolor{violet}{54.07} \\
  PSiLU & 80.54  &\textcolor{violet}{45.45} & 84.73 & \textcolor{violet}{55.20}\\
  %PSSiLU &  & & 84.51 & 56.88 \\
  PSSiLU & 81.85 & \textcolor{violet}{44.70} & \textbf{84.79} & \textbf{\textcolor{violet}{58.21}}\\
\hline

\end{tabular}}

\resizebox{0.55\textwidth}{!}{
\subfloat[WRN-28-10]{
\label{tab:wrn_adv_summary}
\begin{tabular}{ |c|c|c|c|c| } 
\hline
& \multicolumn{2}{|c|}{\textbf{CIFAR-10}} & \multicolumn{2}{|c|}{\textbf{+DDPM-6M}}\\
\hline
Activation & Natural & AA & Natural & AA\\
\hline
  ReLU & 83.39 & 45.98& 85.92 & 59.27\\
  PReLU & 82.75  &43.62 & 86.04  &58.74\\
    \hline

  ELU & 79.66  &45.85 & 81.09  & 50.79\\
  PELU  & 83.32 & 43.85 & 85.83 & 58.90\\
  \hline
  Softplus & 79.99  & 44.41& 78.86 &  49.14\\
  PSoftplus & 82.94 & \textcolor{violet}{46.68} & 86.60 & \textcolor{violet}{60.94}\\
  \hline
  $\pprelu$ &81.71& 45.05 & 86.05 & 59.13\\
  ReBLU &83.16 & \textcolor{violet}{46.93} & 86.39 & \textcolor{violet}{59.62}\\
  \hline
  SiLU & 84.17  & \textcolor{violet}{47.51} &  84.90 & 55.10\\
  PSiLU & 82.41 &\textcolor{violet}{47.03} & 86.47 &  \textcolor{violet}{60.37}\\
  PSSiLU & \textbf{86.02} & \textcolor{violet}{\textbf{48.26}} & \textbf{87.02}  & \textcolor{violet}{\textbf{61.96}} \\
\hline

\end{tabular}}
}}
\caption{Natural and robust accuracy of PGD adversarially trained models of various activation functions with respect to $\ell_{\infty}$ attacks with radius 0.031.  The AA column gives the robust accuracy of attacks generated through AutoAttack on the test set. We highlight robust accuracies larger than ReLU in purple. %\sdcomment{update columns to 2 decimal places}
\vspace{-10pt}
}
\label{tab:summary}
\end{table}

\begin{figure}
    \centering
    \includegraphics[width=\textwidth]{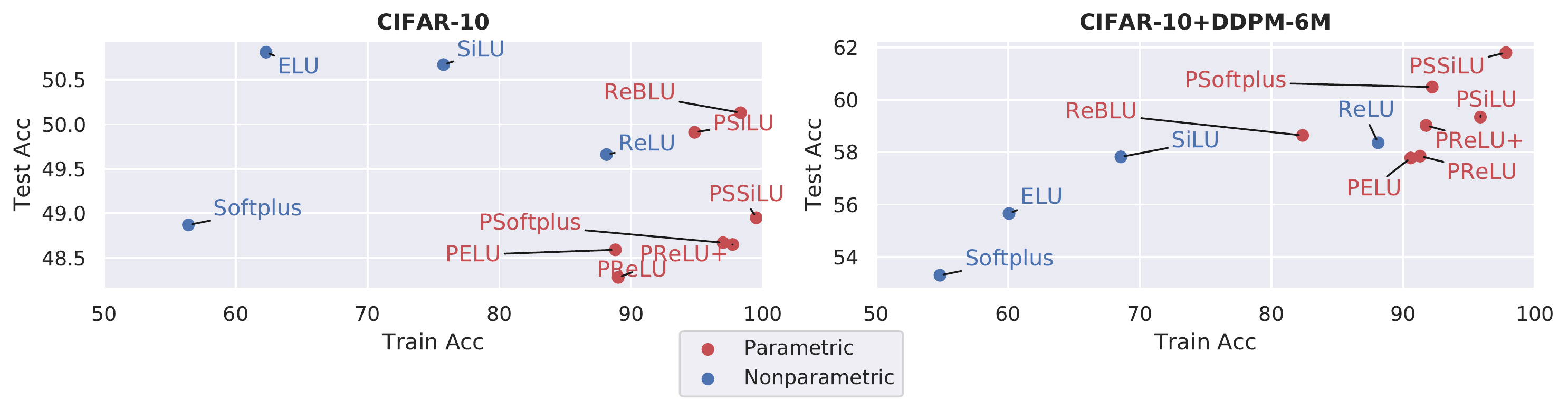}
    \caption{ Highest PGD train and test accuracy for ResNet-18 models.  Train accuracy is measured only on CIFAR-10.}
    \label{fig:train_test}
\end{figure}

To further investigate the impact of additional data on PAFs, we measure the highest train and test robust accuracy using PGD-10 achieved during training. We plot these values in Figure \ref{fig:train_test} for CIFAR-10 and CIFAR-10+DDPM-6M.  PAFs can achieve higher train accuracy compared to nonparametric activation functions, showing that PAFs improve on the expressivity of the model and allow the model to fit to the training set better during adversarial training.  However, we also observe that without the additional DDPM-6M data, PAFs are unable to generalize well to the test set, suggesting that the potential of PAFs is locked behind the use of additional training data.

\subsection{Improving robust performance through regularizing PSSiLU}
The ability of PAFs to quickly overfit to the data suggests that regularization may improve the performance of PAFs.  Unlike the other PAFs, PSSiLU introduces 2 additional parameters into training, which allows it to more easily overfit to the training data compared to other activation functions.  Thus, we regularize the value of $\beta$ in PSSiLU by adding $\lambda |\beta|$ to the loss.  We choose $\lambda = 10$ as the default value and show the effect of varying lambda in Appendix \ref{app:pssilu_reg}.  Results for regularized PSSiLU are displayed in all tables as the PSSiLU entry.

\textbf{Achieving state of the art robust accuracy with PSSiLU and DDPM-6M.}
%We observe that after regularizing the $\beta$ parameter, the performance of PSSiLU increases by 1.33\% on ResNet-18, which validates the importance of regularizing parameters of PAFs to improve robustness.  Additionally, w
We observe that for ResNet-18 and WRN-28-10, PSSiLU achieves both high clean and high robust accuracy.  Compared to ReLU, we observe that PSSiLU improves robust performance by a total of 4.54\% \textit{while only adding 2 parameters into the network architecture}.  This accuracy is only 1.06\% lower than the result for WRN-28-10 model with ReLU activations in Table \ref{tab:wrn_adv_summary}, but WRN-28-10 has 25.3M more parameters than ResNet-18. In other words, we bridged a 5.60\% performance gap produced by 25.3M additional parameters by 4.54\% by adding only 2 parameters into the architecture. Moreover, with the additional DDPM-6M data on ResNet-18, PSSiLU improves over the robust performance of SiLU by 4.14\% and PSiLU by 3.01\%, both of which can be modeled by PSSiLU.

On WRN-28-10, PSSiLU achieves 87.02\% clean accuracy and 61.96\% robust accuracy, improving on clean accuracy by 1.10\% and robust accuracy by 2.69\% over ReLU, making our WRN-28-10 model the best performing in its category on RobustBench \citep{croce2020reliable}.

\textbf{Consistency of the PSSiLU family.} The function class of PSSiLU captures that of PSiLU and SiLU.  PSiLU can be thought of as PSSiLU with $\beta = 0$, which can be achieved by placing a large regularization term on $|\beta|$.  Similarily, SiLU is PSSiLU with $\alpha = 1$ and $\beta = 0$, and can be achieved by PSSiLU with a large regularization term on $|\alpha - 1|$ and $|\beta|$.  Across datasets and architectures tested, we find that a member of the PSSiLU family is able to consistently obtain high robust accuracy.  We also find that this also generalizes to TRADES adversarial training, $\ell_2$ attacks, and other sources of additional data (Appendix \ref{app:additional_adv}).  This suggests that the function class of captured by PSSiLU works well in conjunction with adversarial training in improving robustness.

\textbf{Consistency of smooth PAFs.} We find that smooth PAFs (PSoftplus, PSiLU, and PSSiLU) often \textit{improve robust accuracy over ReLU} even when additional data is not present.  In Appendix \ref{app:additional_adv}, we find that this pattern holds across datasets and architectures and generalizes to TRADES adversarial training and $\ell_2$ attacks.

\subsection{Visualizing Learned Shapes of Parametric Activation Functions}
Previously, in Section \ref{sec:directeffect}, we searched for a parameterization of activation function shape that controls factors which directly influence robustness independently of adversarial training.  This begs the question, how are the properties observed (positive outputs on negative inputs, high curvature) related to shapes learned through adversarial training?  In this section, we visualize the shapes of PAFs learned through the adversarial training objective.

\begin{figure*}[h]
    \center
    \includegraphics[width=\textwidth]{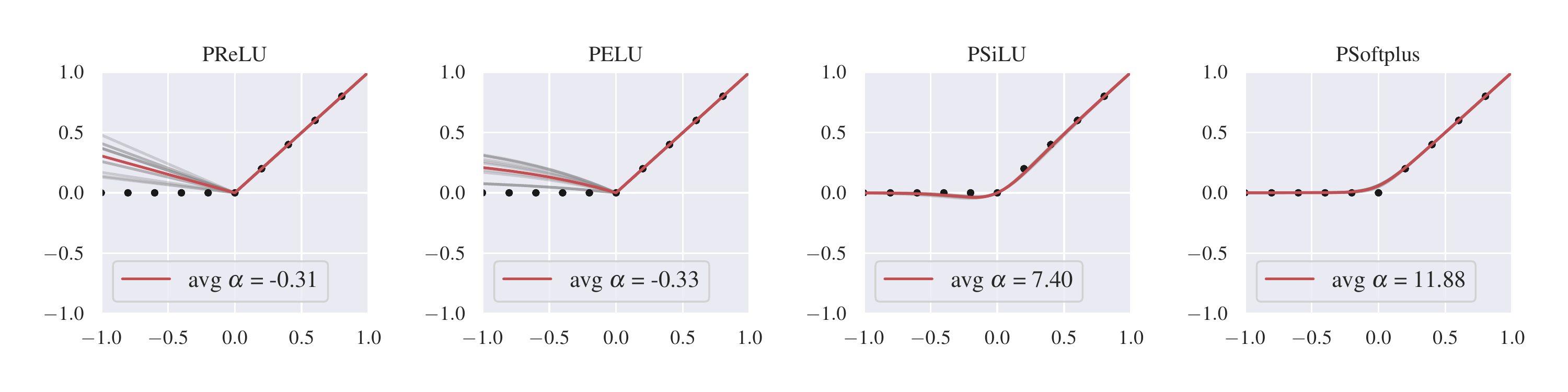}
    \caption{Learned shapes of PAFs across 11 models of various architectures (WRN-28-10, ResNet-18, VGG) trained using PGD adversarial training on various datasets (CIFAR-10, CIFAR-100, ImageNette).  Each grey line represents the shape learned by a single model.  The red line represents the average of the learned $\alpha$s across all models.  The dotted black line represents ReLU.}
    \label{fig:learned_adv_shape_all}
\end{figure*}

\begin{wrapfigure}{r}{0.4\textwidth}
    \vspace{-10pt}

    \centering
    \includegraphics[width=0.4\textwidth]{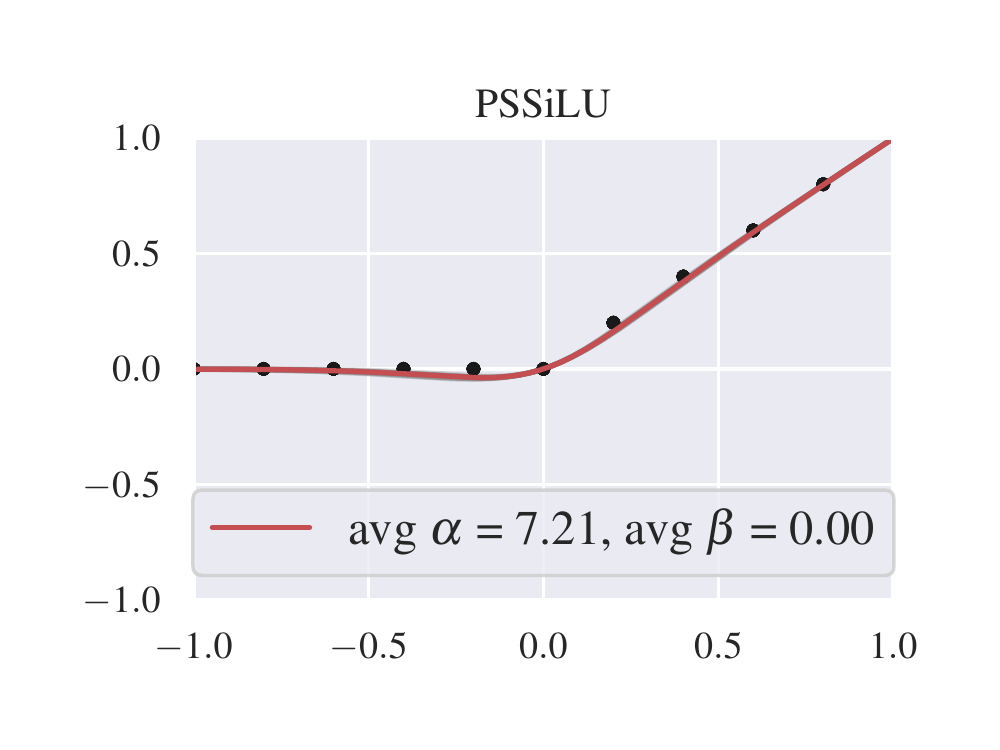}
        \vspace{-25pt}

    \caption{Learned shapes of activation functions across 11 models of various architectures (WRN-28-10, ResNet-18, VGG) trained using PGD adversarial training on various datasets (CIFAR-10, CIFAR-100, ImageNette).  Each grey line represents the shape learned by a single model.  The red line represents the average of the learned $\alpha$s and $\beta$s across all models.  The dotted black line represents ReLU.}
    \vspace{-40pt}
   \label{fig:pssilu_shape}

\end{wrapfigure}

We present the learned shapes of PReLU, PELU, PSiLU and PSoftplus in Figure \ref{fig:learned_adv_shape_all}.  Additionally, we present the learned shape of PSSiLU in Figure \ref{fig:pssilu_shape}.

\textbf{Relation to Trends from Section \ref{sec:directeffect}.}
We observe that across architectures and datasets, PAFs optimize towards the same qualities that were found to improve robustness for standard trained models: positive behavior on negative inputs and high curvature.  In Figure \ref{fig:learned_adv_shape_all}, we can see that for all models, PReLU and PELU optimize to give positive output on negative inputs while PSiLU and PSoftplus both optimize towards the shape of ReLU. This suggests that these patterns in shape observed can also help the model better fit the training data.
%\paragraph{Comparison to shapes learned through standard training} For each model trained with adversarial training, we also train a corresponding model using standard training.  We plot the average shape learned by standard training in blue in Figure \ref{fig:learned_adv_shape_all} and \sdcomment{PSSiLU without regularization figure ref}. We observe that the shapes learned through adversarial training are also correlated with those learned through standard training. In comparison to standard training, the parameters of standard trained models optimize towards larger positive outputs on negative inputs on PReLU and PELU compared to adversarial training.  Additionally, adversarial training optimizes towards higher curvature on PSiLU and PSoftplus than standard training. \sdcomment{add comment about PSSiLU no regularization as well}

\textbf{Optimization itself improves robust performance.} In Figure \ref{fig:pssilu_shape}, we visualize the shapes learned by PSSiLU with regularized $\beta$.  Surprisingly, we find that the $\beta$ parameters optimize to be 0, which reduces PSSiLU to PSiLU.  However, compared to PSiLU, we observed from Table \ref{tab:summary} that with additional DDPM data, regularized PSSiLU significantly improves robustness over PSiLU.  Specifically, PSSiLU improves performance over PSiLU by 3.01\% on ResNet-18 and by 1.59\% on WRN-28-10.  These results suggest that in addition to increasing the size of the function class, adding parameters  can also help with optimization, allowing adversarial training to converge to a better minimum.
\vspace{-10pt}

\section{Limitations and Future Directions}
%Our study on the impact of activation functions on robustness opens several directions for future research.
In our work, we show that by parameterizing activation functions in DNNs, we can improve robust accuracy obtained through adversarial training.  However, we observe two challenges with using PAFs: the necessity of additional data and regularization of PAF parameters.

%\textbf{Tuning regularization of parameters}  Choosing an appropriate amount of regularization on PAF parameters can improve the performance of adversarial training with PAFs.  However, hyperparameter tuning on adversarially trained models can be difficult due to long training time, especially on larger models and when using additional data during training.  In future work, we aim to find a way to efficiently tune regularization on parameters.  Since this work did not perform hyperparameter tuning, additional research on this front may further boost performance of PSSiLU and other PAFs.
\textbf{Sources of additional data.} We find that using additional data boosts the performance of adversarial training with PAFs.  Previous works have indicated that on larger datasets such as ImageNet, current generative models are unable to provide samples that improve the performance of adversarial training \citep{sehwag2021improving}.  Additional progress on generative models would also improve the performance of adversarial training with PAFs on larger datasets.

\textbf{Regularizing Parameters.} When additional data is not present, we observed that PAFs do not consistently outperform their nonparametric counterparts.  This result motivates the use of regularization.  In this work, we use L1 regularization on the $\beta$ parameter of PSSiLU.  A future direction of this work is to explore other forms of regularization such as curvature regularization or Lipschitz constant regularization since these properties of DNNs have been shown to be related to robustness \citep{singla2021low, moosavi2019robustness,NEURIPS2019_0defd533, pauli2021training}.

\textbf{Combining PSSiLU with other training pipelines.}  Our training pipeline is based off of \citet{sehwag2021improving}'s pipeline for PGD training with additional synthetic data.  Another future direction of this work is combining PSSiLU with other training pipelines which achieve higher robust performance on RobustBench \citep{croce2020robustbench} to see if we can further improve robust performance.

%\paragraph{Tuning the parameter of PSiLU}  Across CIFAR-10 and ImageNette datasets, we found that one of PSiLU or SiLU is able to outperform ReLU networks by up to 2\% robust accuracy.  In some cases, PSiLU performs worse than ReLU while SiLU outperforms ReLU by a margin while in other cases SiLU performs worse than ReLU while PSiLU is able to outperform ReLU.  Since SiLU is PSiLU with parameter $\alpha=1$, this finding raises the question, how can we better tune the parameter of PSiLU so that we can maintain good performance throughout all model and dataset configurations?

\section{Conclusion}
In this work, we study the impact of parameterizing activation functions on robustness through adversarial training.  We identify qualities in activation function shape that improve robustness and combine these properties into a new PAF (PSSiLU). We combine learnable PAFs with adversarial training and find that by introducing as many as 1-2 additional parameters into the network architecture, PAFs can significantly improve robustness over ReLU.  Overall, this work demonstrates the potential of using learnable PAFs for enhancing robustness of machine learning against adversarial examples.

\section{Acknowledgements}
We would like to thank Vikash Sehwag and Chong Xiang for their discussions on this project and feedback on the paper draft.  This work was supported in part by the National Science Foundation under grants CNS-1553437 and CNS-1704105, the ARL’s Army Artificial Intelligence Innovation Institute (A2I2), the Office of Naval Research Young Investigator Award, Schmidt DataX award, and Princeton E-ffiliates Award.  This material is based upon work supported by the National Science Foundation Graduate Research Fellowship under Grant No. DGE-2039656. Any opinions, findings, and conclusions or recommendations expressed in this material are those of the author(s) and do not necessarily reflect the views of the National Science Foundation.

\appendix
\section{Experimental Setup Details}
\label{app:experimental_setup}
\subsection{Standard Training Experimental Setup Additional Details}
\label{app:standardsetup}
\paragraph{Models.} We train ResNet-18 \citep{he2016deep}, WideResNet-28-10 \citep{zagoruyko2016wide}, and VGG-16 \citep{simonyan2015vgg} models.  For each model, we replace ReLU activations with activation functions described in Section \ref{sec:act_def}.  For each activation function, we train multiple models, setting the value of the parameter to those shown in Figure \ref{fig:act_plots_all}.
\paragraph{Datasets.}  We train ResNet-18, WRN-28-10, and VGG-16 models on CIFAR-10  \citep{krizhevsky2009learning}.  For ResNet-18 models, we also perform experiments on ImageNette (a 10 class subset of ImageNet) \citep{howardimagenette} and CIFAR-100 \citep{krizhevsky2009learning}.  For ImageNette, we resize images to 224x224 before passing the images into ResNet-18.
\paragraph{Training Setup}  We train models using SGD with initial learning rate of 0.1 and use cosine annealing learning rate scheduling \citep{loshchilov2016sgdr}. We train all models for 100 epochs and perform evaluation on the model saved at the epoch which has the highest accuracy on the test set.  For ResNet-18 models on CIFAR-10, we run 3 trials.  For each trial, all models are seeded to the same seed.  For all other models we run a single trial.
\paragraph{Evaluation Setup.} For measuring adversarial sensitivity, we consider the adversary constrained to an L-infinity budget.  To estimate the smallest perturbation radius for misclassification, we perform a binary search on radius size for 4-step PGD \citep{madry2017towards} with step size of 0.0078.  We use 4-step PGD to increase the efficiency of the binary search algorithm.  For query-restricted black-box adversary, we use Square attack \citep{andriushchenko2020square, croce2020reliable} with budget $\epsilon=0.031$ and compute the robust accuracy on adversarial examples found within 1000 queries. We measure square attack robust accuracy and PGD radius over images in the test set that are initially classified correctly by the model.

\subsection{Adversarial Training Experimental Setup}
\label{app:adversarialsetup}
\paragraph{Models.} We train ResNet-18 \citep{he2016deep}, WRN-28-10 \citep{zagoruyko2016wide}, and VGG-16 \citep{simonyan2015vgg} models.  For each activation function tested, we replace all ReLU within both models with that activation function.  We test parametric activations described in Section \ref{sec:act_def} and allow the parameter $\alpha$ (and $\beta$ for PSSiLU) to be optimized through training.  Additionally, we test nonparametric variants of these activation functions: ReLU, ELU, SiLU, and Softplus.  We initialize all parametric activation functions to the shapes of their nonparametric variants (ReLU for PReLU and PBLU, ELU for PELU, SiLU for PSiLU and PSSiLU, and Softplus for PSoftplus).  For all parametric activation functions, we share the parameter across activations within the network, so PSSiLU adds 2 parameters to the network overall while all other parametric activation functions add 1 parameter.

\paragraph{Datasets.}  Overall, we experiment with 3 datasets: CIFAR-10 \citep{krizhevsky2009learning}, ImageNette \citep{howardimagenette}, and CIFAR-100 \citep{krizhevsky2009learning}.  Additionally, for CIFAR-10 experiments, we consider the setting of training with and without additional data.  We consider 2 sources of additional data for CIFAR-10 models: DDPM-6M \citep{sehwag2021improving} and TI-500K \citep{carmon2019unlabeled}.  DDPM-6M is a synthetic dataset of 6 million images generated by DDPM, a generative model \cite{ho2020denoising}. Previous works have shown that using samples from DDPM can improve robustness through adversarial training \citep{sehwag2021improving, rebuffi2021fixing}.  TI-500K is a subset of TinyImages which matches the distribution of CIFAR-10 and has also been shown to improve robustness when used in adversarial training \citep{carmon2019unlabeled}.  We train ResNet-18 models on CIFAR-10, ImageNette, and CIFAR-10+DDPM. For ImageNette, we resize images to 224x224 before passing the images into ResNet-18.  We train WRN-28-10 and VGG models on CIFAR-10, CIFAR-10+DDPM, and CIFAR-10+TI-500k.  We also train WRN-28-10 models on CIFAR-100.

\paragraph{Training Details.}  For the bulk of experiments, we use PGD adversarial training \citep{madry2017towards} and train models for 200 epochs.  We also train a set of ResNet-18 models on CIFAR-10 and L-infinity adversary with TRADES adversarial training \cite{zhang2019theoretically} with $\beta = 0.6$ (Appendix \ref{app:trades}.  For all architectures, we train with 10 step PGD with L-infinity budget of 0.031 and step size of 0.0078.  For WRN-28-10 models on CIFAR-10, we also train with 10-step PGD with L-2 budget of 0.5 and step size of 0.075.  We train models with SGD with learning rate 0.1 and cosine annealing learning rate scheduling.  We seed all models to 12345 to control for differences caused by randomness in initialization.  For PSSiLU models, we apply regularization on the magnitude of parameter $\beta$ to restrict the slope on negative inputs so that it remains small.  Specifically, for PSSiLU models, we add a $\lambda |\beta|$ term to the loss function where $\lambda = 10$.  Additionally, since a small fluctuation of $\beta$ leads to a large change in activation function shape, we clip the gradients of the $\beta$ parameter to have norm 0.01.

\paragraph{Evaluation Details.} We evaluate models saved at the epoch which had the highest PGD adversarial accuracy on the test set.  We perform our final evaluation of robustness using AutoAttack \citep{croce2020reliable}.  During evaluation, we use the same adversarial budget that was used to train the model.

\section{Additional Results for Standard Trained Models}
\subsection{Clean Accuracies of Standard Trained Models}
We report the minimum and maximum classification accuracies for each model and dataset combination across all activations and parameter values in Table \ref{tab:accrange}
\begin{table}[h]
    \centering
    \begin{tabular}{|c|c|c|c|}
    \hline 
         Model & Dataset & Min Acc & Max Acc \\
         \hline
         ResNet-18 & CIFAR-10 & 89.1 & 95.3 \\
         WRN-28-10 & CIFAR-10 & 89.2 & 96.0 \\
         VGG-16 & CIFAR-10 & 76.4 & 94.1 \\
         ResNet-18 & Imagenette & 79.4 & 92.7 \\
         ResNet-18 & CIFAR-100 & 70.2 & 77.8 \\
         \hline
    \end{tabular}
    \caption{Minimum and maximum values for clean accuracy of standard trained models across all activations and parameter values tested.}
    \label{tab:accrange}
\end{table}
\subsection{Generalization of observed trends}
\label{app:generalization}
\begin{figure}
    \centering
    \includegraphics[width=0.9\textwidth]{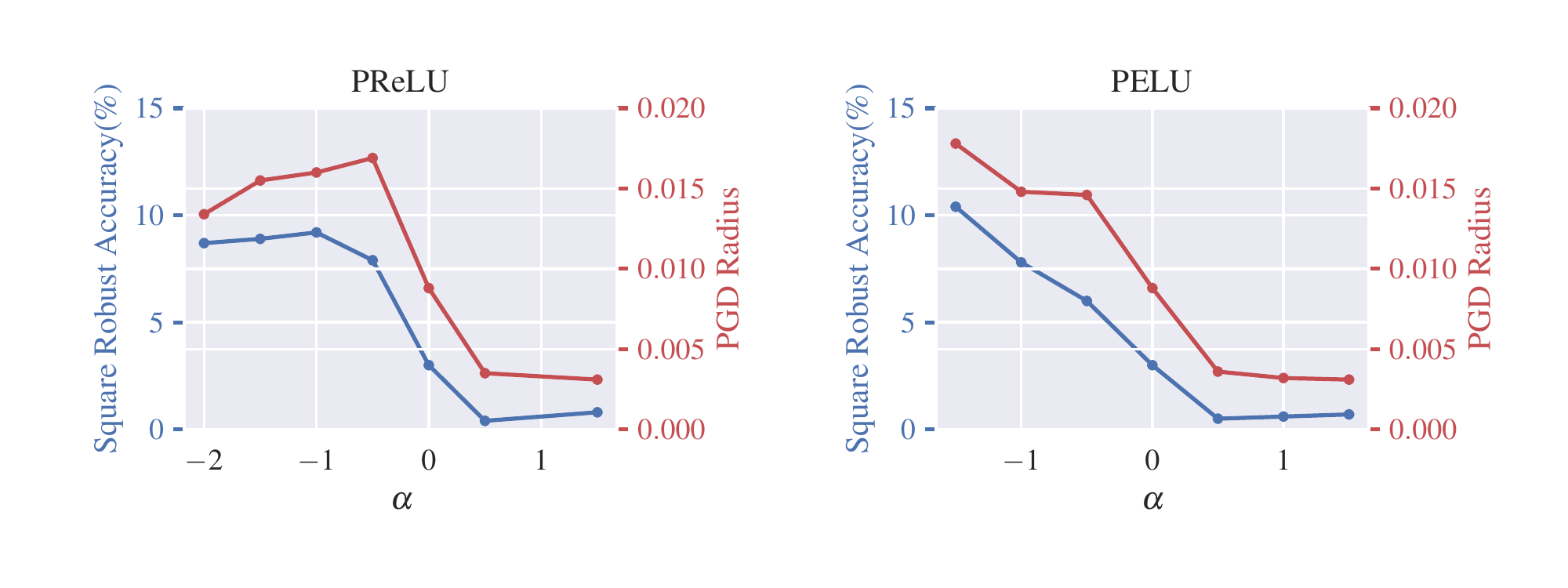}
    \includegraphics[width=0.9\textwidth]{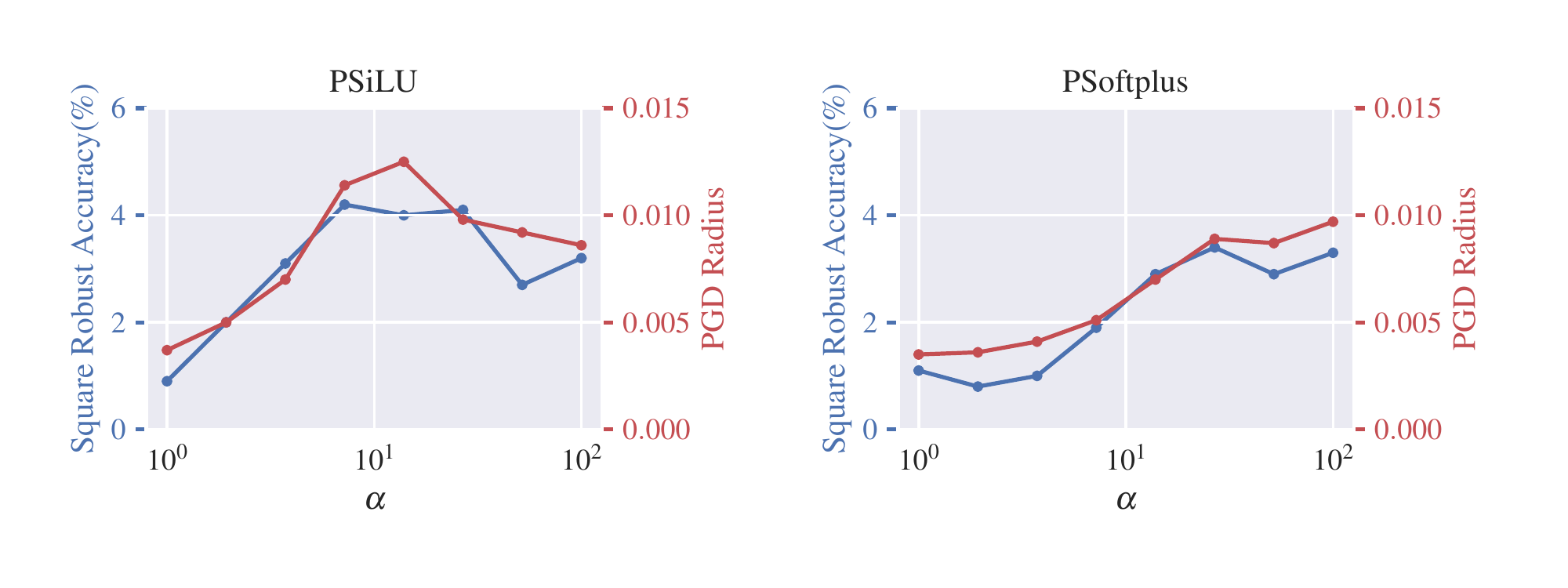}
    \caption{Square robust accuracy and average minimum PGD radius for WRN-28-10 models trained on CIFAR-10 with various parameter $\alpha$.}
    \label{fig:prelu_pelu_standard_wrn}
\end{figure}

\begin{figure}
    \centering
    \includegraphics[width=0.9\textwidth]{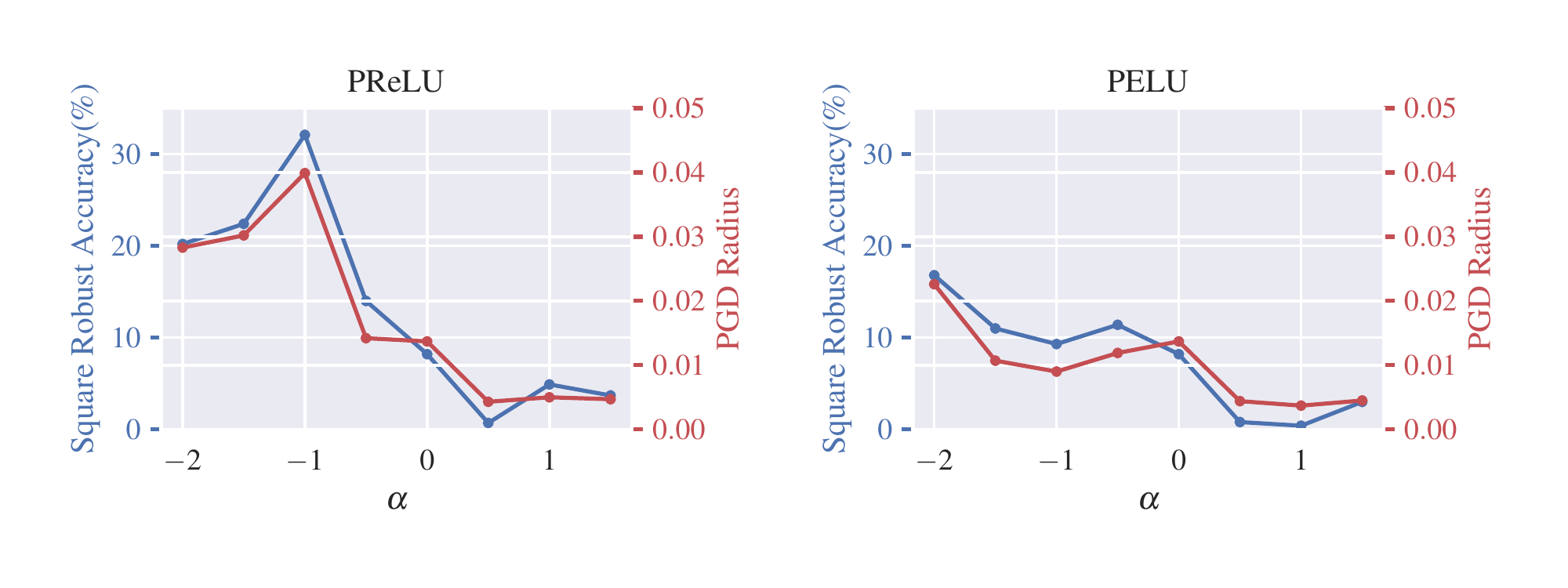}
    \includegraphics[width=0.9\textwidth]{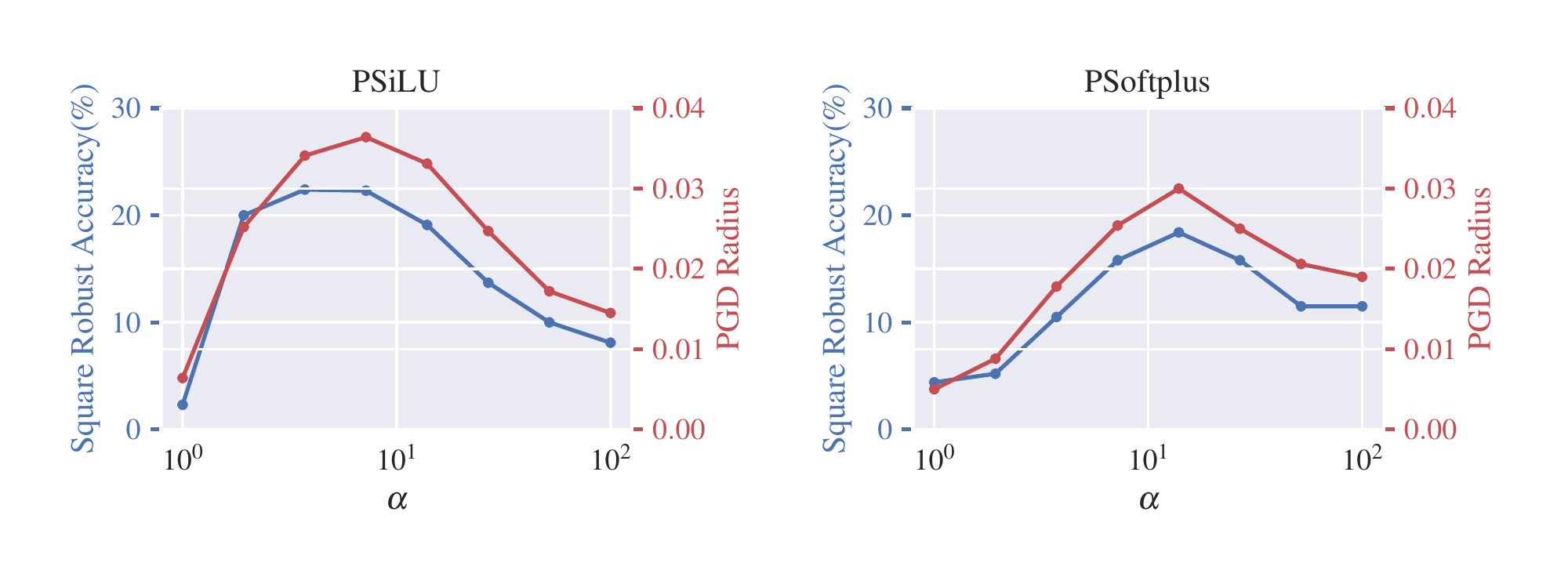}
    \caption{Square robust accuracy and average minimum PGD radius for VGG-16 models trained on CIFAR-10 with various parameter $\alpha$.}
    \label{fig:prelu_pelu_standard_vgg}
\end{figure}

To test whether the trends seen in Section \ref{sec:directeffect} generalize to other model architectures, we repeat experiments on WRN-28-10 (Figure \ref{fig:prelu_pelu_standard_wrn}) and VGG-16 (Figure \ref{fig:prelu_pelu_standard_vgg}).  We find that across architectures, the trends for PReLU, PELU, PSiLU, and PSoftplus are consistent.

\begin{figure}
    \centering
    \includegraphics[width=0.9\textwidth]{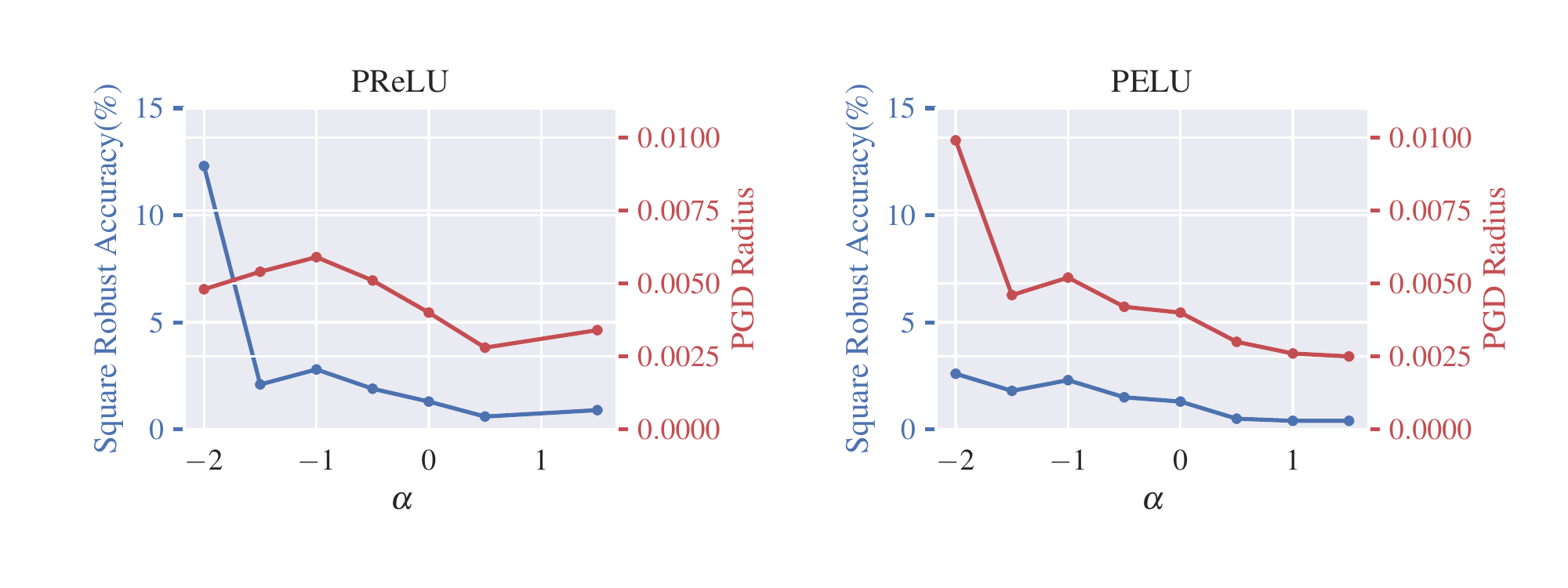}
    \includegraphics[width=0.9\textwidth]{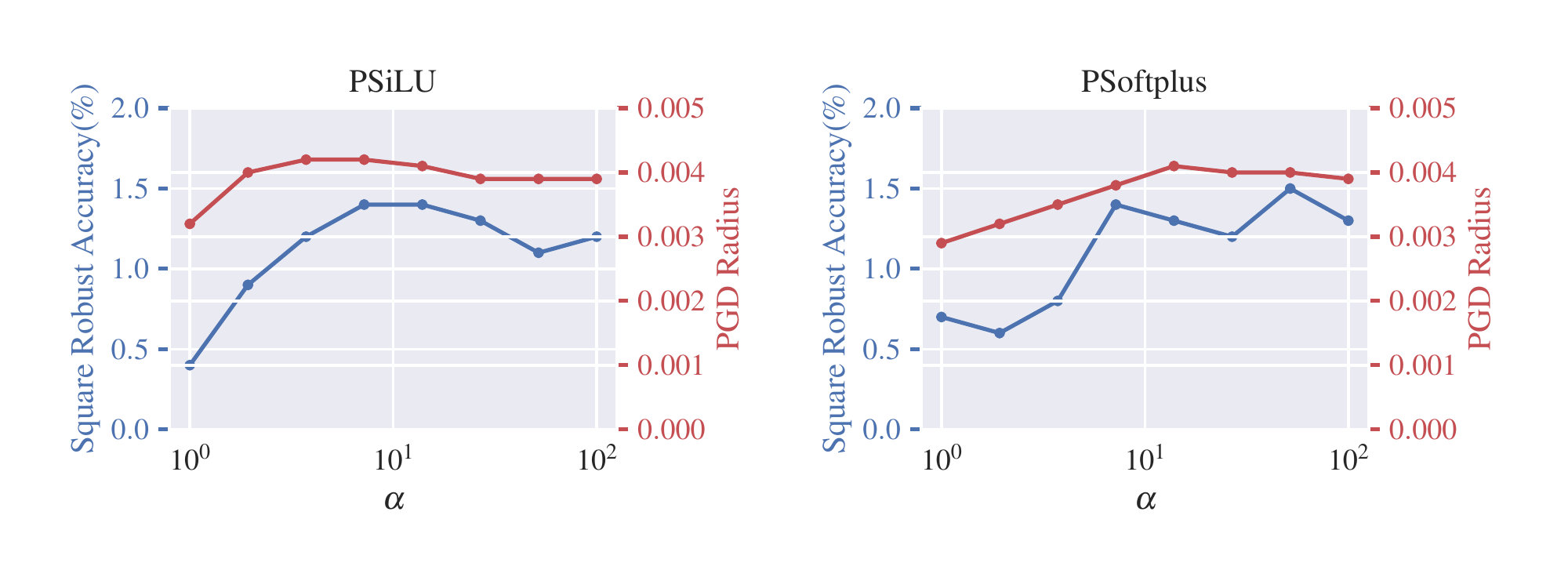}
    \caption{Square robust accuracy and average minimum PGD radius for ResNet-18 models trained on CIFAR-100 with various parameter $\alpha$.}
    \label{fig:prelu_pelu_standard_cif100}
\end{figure}

\begin{figure}
    \centering
    \includegraphics[width=0.9\textwidth]{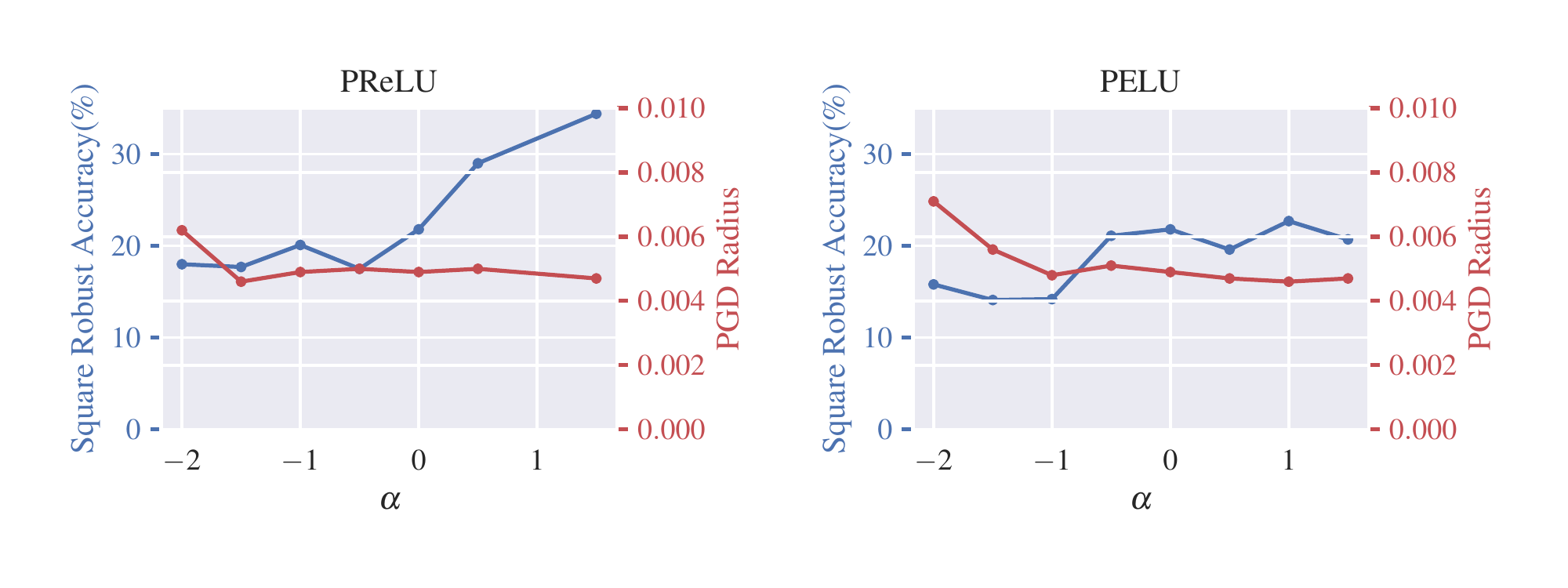}
    \includegraphics[width=0.9\textwidth]{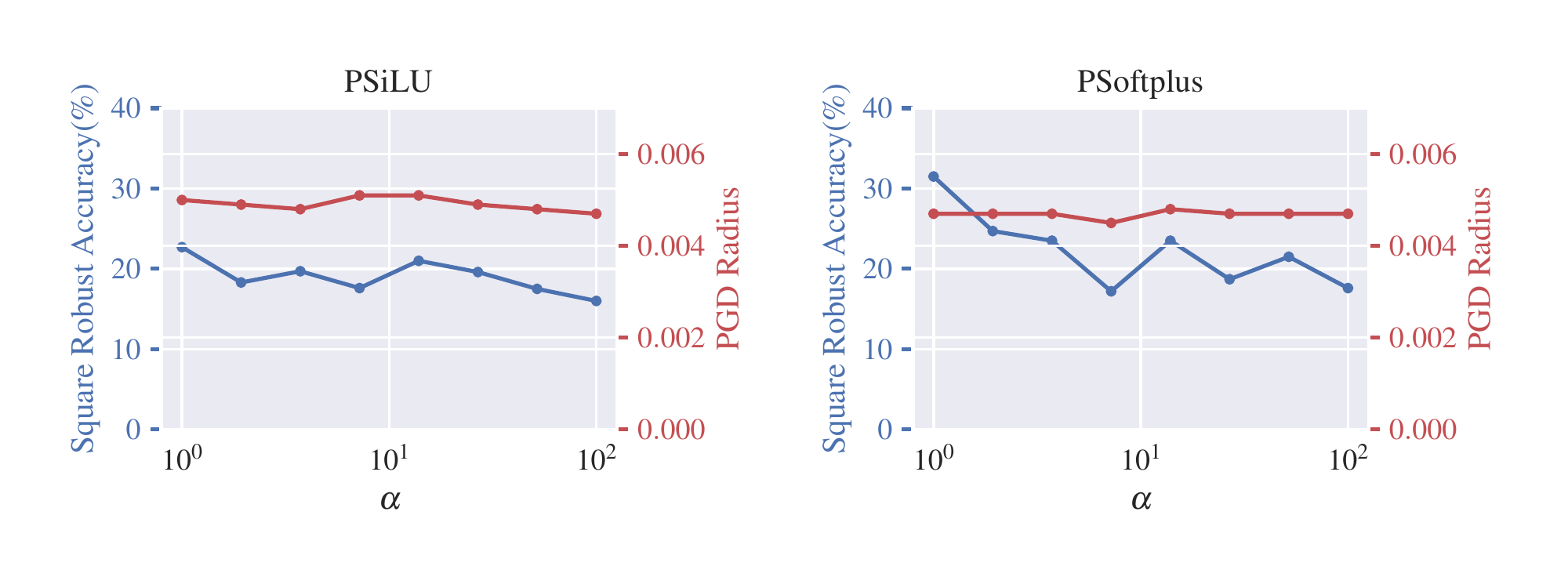}
    \caption{Square robust accuracy and average minimum PGD radius for ResNet-18 models trained on ImageNette with various parameter $\alpha$.}
    \label{fig:prelu_pelu_standard_imagenette}
\end{figure}

To test whether these patterns also generalize across dataset, we repeat experiments on CIFAR-100 (Figure \ref{fig:prelu_pelu_standard_cif100}) and ImageNette (Figure \ref{fig:prelu_pelu_standard_imagenette}).  We find that these trends are clear for CIFAR-100 but are not clear in ImageNette.  For ImageNette, we find that there is little variation across PGD radius and Square robust accuracy is not always correlated with the measured radius as is observed for CIFAR-10 and CIFAR-100.

\subsection{Behavior for activation functions varying in the positive regions}
\label{app:posbehavior}
\begin{figure}
    \centering
    \includegraphics[width=0.9\textwidth]{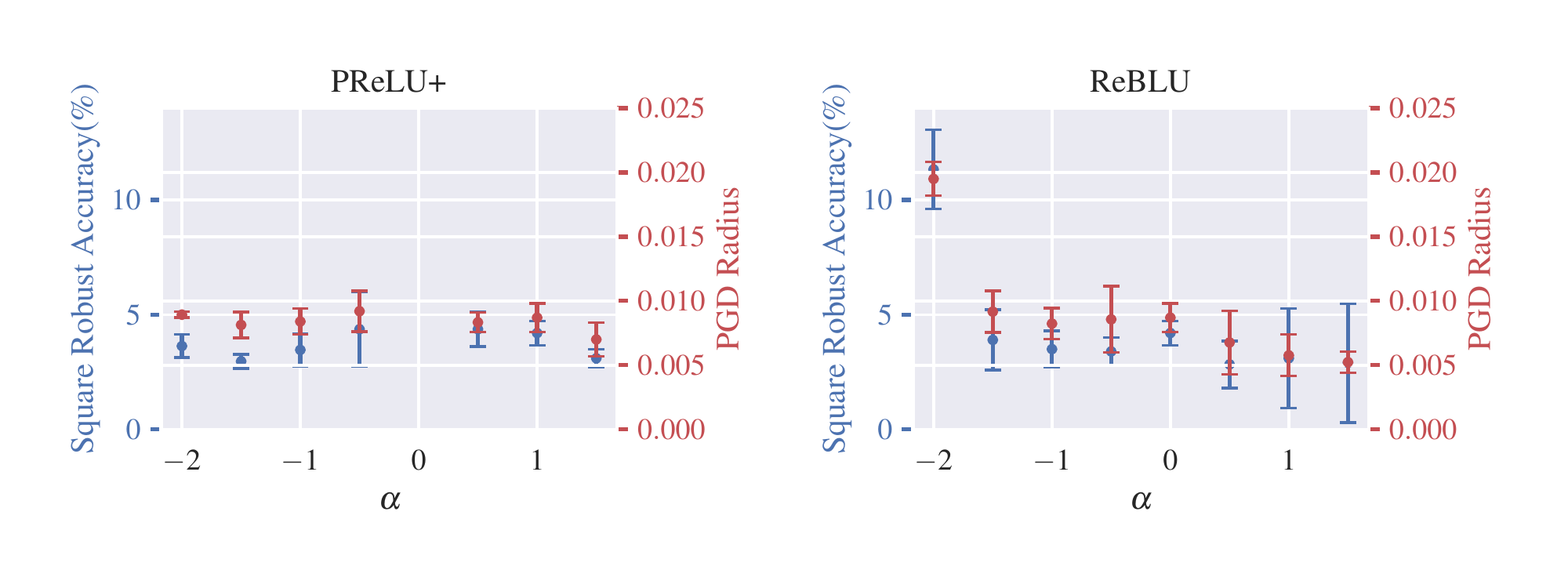}
    \caption{Square robust accuracy and average minimum PGD radius for ResNet-18 models trained on CIFAR-10 with various parameter $\alpha$ for $\pprelu$ and PBLU activations.  Errors bars are computed over 3 trials.}
    \label{fig:pprelu_pblu_standard}
\end{figure}

In addition to testing activations with varied behavior on negative inputs and around 0, we also measure the square robust accuracy and average minimum PGD radius for $\pprelu$ and PBLU.  For these activation functions, we were unable to observe any clear trends for the impact of positive behavior.  We provide the plot for ResNet-18 models trained on CIFAR-10 in Figure \ref{fig:pprelu_pblu_standard}.  We find that PPReLU exhibits high variance in behavior making trends unclear.  Figure \ref{fig:pprelu_pblu_standard} suggests that $\alpha < 0$ may lead to higher perturbation stability, we find that this is inconsistent across architectures.  For instance, we observe the opposite trend in Figure \ref{fig:prelu_pelu_standard_vgg} which shows the behavior of PPReLU and PBLU for VGG-16 models trained on CIFAR-10.  There are no consistent trends across these activation functions, which suggests that positive behavior is less important for robustness.  Thus, we do not introduce a parameter to control positive behavior on PSSiLU.

\begin{figure}
    \centering
    \includegraphics[width=0.9\textwidth]{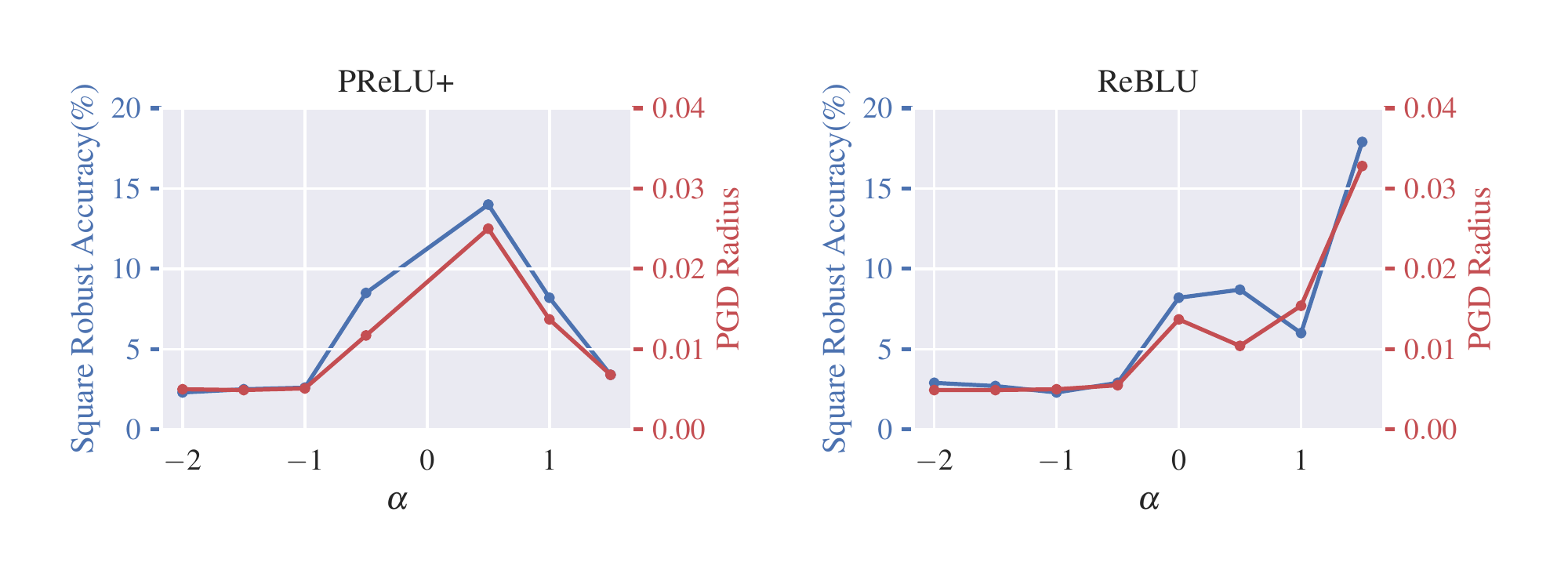}
    \caption{Square robust accuracy and average minimum PGD radius for VGG-16 models trained CIFAR-10 with various parameter $\alpha$ for $\pprelu$ and PBLU activations.}
    \label{fig:pprelu_pblu_standard_vgg}
\end{figure}

\subsection{Empirical Lipschitz constant of PReLU and PeLU Models}
\label{app:lipschitz}
\begin{figure}
    \centering
    \includegraphics[width=\textwidth]{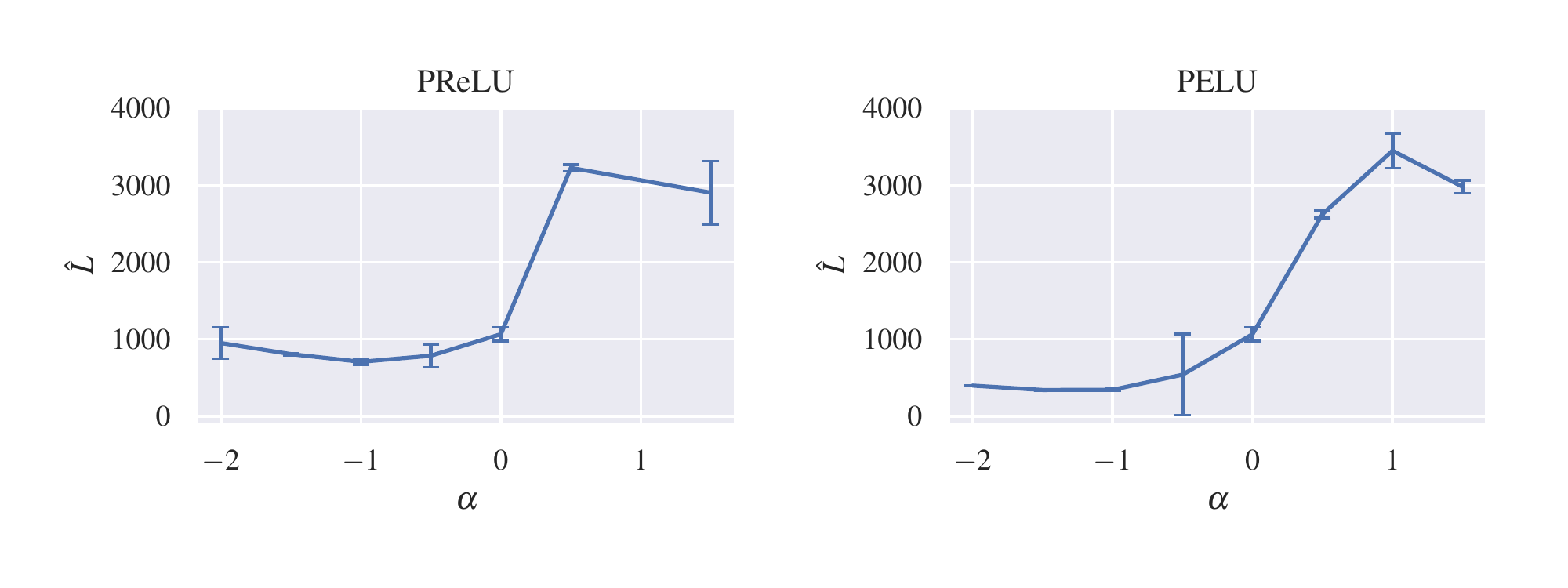}
    \caption{Empirical Lipschitz constant of PReLU and PELU ResNet-18 models trained on CIFAR-10 at varied value of parameter $\alpha$.  Lower empirical Lipschitz constant suggests that the model outputs are more stable in the presence of perturbations.  The trend in Empirical Lipschitz constant matches the trends observed in PGD radius and Square robust accuracy.}
    \label{fig:lipschitz}
\end{figure}

For PReLU and PELU, we observe that for $\alpha < 0$ when the magnitude of $\alpha$ becomes large, the adversarial difficulty decreases.  We hypothesize that this trend is due to neural network Lipschitz constant.  When $|\alpha|$ grows, the Lipschitz constant for PReLU and PELU also increases. The Lipschitz constant of neural network controls the amount of change that can occur in the output when an input is perturbed, so restricting the magnitude of the Lipschitz constant can improve adversarial robustness \citep{NEURIPS2019_0defd533, jordan2020exactly, pauli2021training}.  Neural network Lipschitz constant depends on the Lipschitz constant of activation functions and weight matrices within the network.  We hypothesize that as $|\alpha|$ becomes large, the Lipschitz constant of the neural network increases due to the increase in Lipschitz constant of the activation function.  To test this, we measure the empirical Lipschitz constant of PReLU and PELU models, where the empirical Lipschitz constant of a model is defined as \citep{yang2020closer}
\begin{equation}
    \hat{L} = \frac{1}{n}\sum_{i=1}^n \max_{\hat{x}_i \in B(\epsilon, x_i)} \frac{||f(x_i) - f(\hat{x}_i)||_1}{||x_i - \hat{x}_i||_{\infty}}
\end{equation}
where $f$ is the model, $\{x_i\}_{i=1}^n$ represent the data inputs and $B(\epsilon, x_i)$ represents a ball of radius $\epsilon$ around $x_i$.  $\hat{x_i}$ can be generated by an adversarial attack.  We compute this quantity using PGD-10 with radius 0.031 and step size 0.0078 to generate $\hat{x}_i$.  The trends for the empirical Lipschitz constant of PReLU and PELU ResNet-18 models is shown in Figure \ref{fig:lipschitz}.  We find that the trends for empirical Lipschitz constant are also consistent with the trends for Square robust accuracy and PGD radius.

\subsection{Generalization of trends for PSSiLU}
\label{app:pssilu_gen}
\begin{figure}[h]
\subfloat[\centering WRN-28-10]{
    \includegraphics[width=0.5\textwidth]{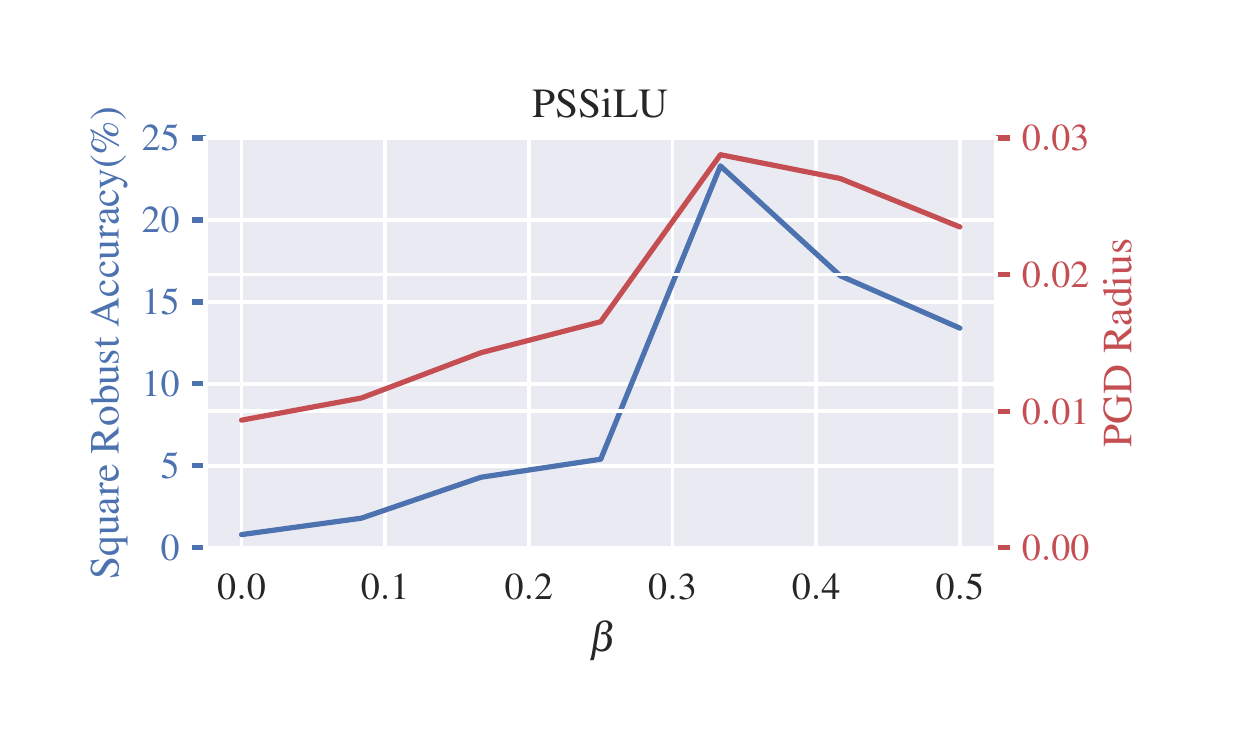} }
    \subfloat[\centering VGG-16]{
    \includegraphics[width=0.5\textwidth]{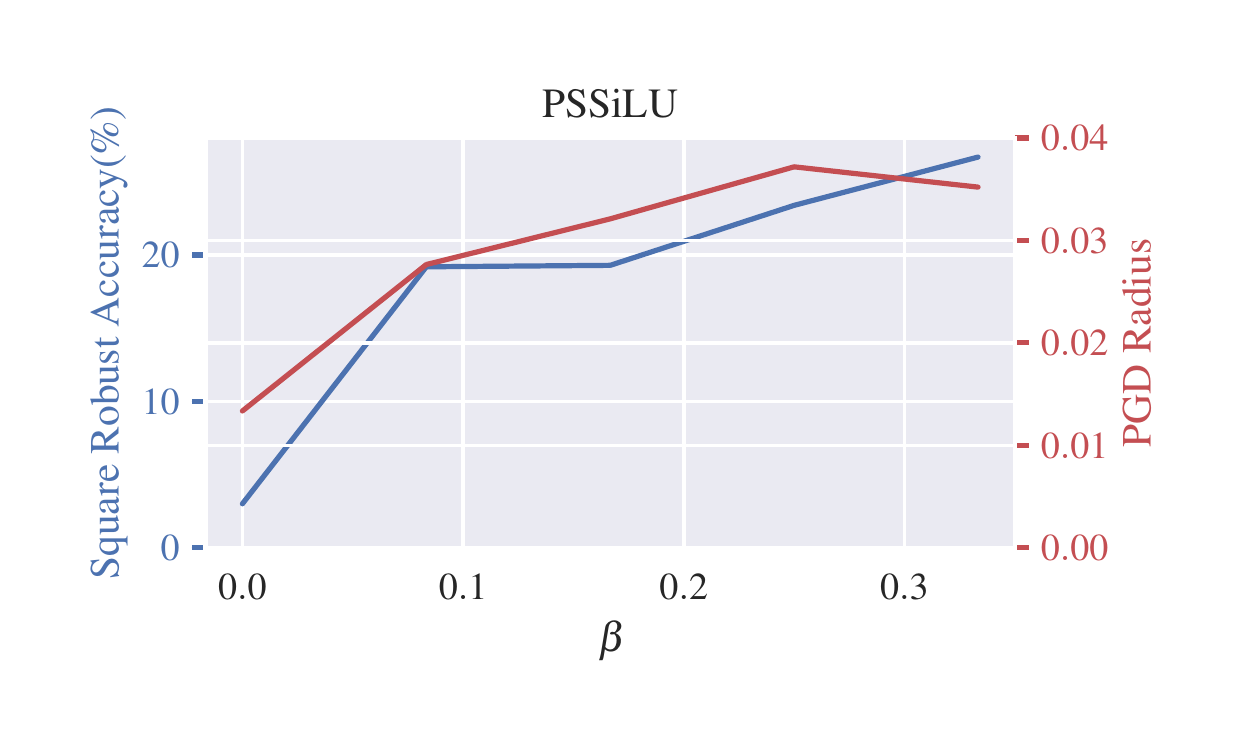}}
    \caption{Square robust accuracy and average minimum PGD radius for PSSiLU across parameter $\beta$ for standard trained WRN-28-10 and VGG-16 architectures.}
    \label{fig:pssilu_beta_arch}

\end{figure}

\begin{figure}[h]
    \centering
    \includegraphics[width=0.5\textwidth]{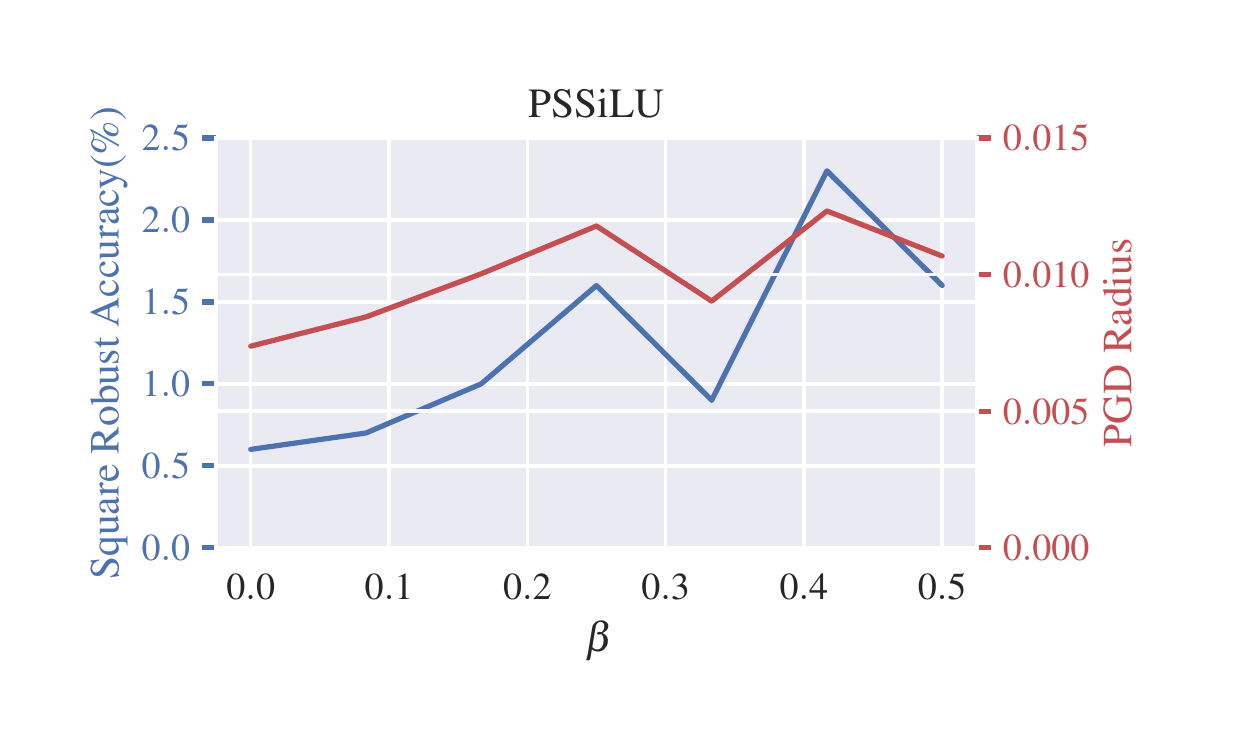}
    \caption{Square robust accuracy and average minimum PGD radius for PSSiLU across parameter $\beta$ for standard trained ResNet-18 models on CIFAR-100.}
    \label{fig:pssilu_beta_cif100}
\end{figure}

To show that the patterns observed on PSSiLU at varied parameter $\beta$ are consistent across architecture and dataset, we report results for Square robust accuracy and PGD radius on WRN-28-10 and VGG-16 architectures in Figure \ref{fig:pssilu_beta_arch}.  Additionally, we report results on CIFAR-100 in Figure \ref{fig:pssilu_beta_cif100}.  We find that as $\beta$ increases the model robustness also increases.  This is consistent with our findings on ResNet-18 models trained on CIFAR-10.

\section{Additional Results for Adversarially Trained Models}
\label{app:additional_adv}
\subsection{Results on VGG-16}
We report additional results for VGG-16 models trained on CIFAR-10 in Table \ref{tab:vgg_adv_summary}.  We find that trends in VGG-16 are generally consistent with those for ResNet-18 and WRN-28-10: the best performing models are within the PSSiLU family and parametric activation functions outperform nonparametric activation functions when there is additional data in training.  Additionally, we find that smooth PAFs (PSoftplus, PSiLU, PSSiLU) improve robust accuracy over ReLU, even without extra data from DDPM-6M.
\begin{table}[ht]
\centering
    \begin{center}
\begin{tabular}{ |c|c|c|c|c| } 
\hline
& \multicolumn{2}{|c|}{\textbf{CIFAR-10}} & \multicolumn{2}{|c|}{\textbf{+DDPM-6M}}\\
\hline
Activation & Natural & AA & Natural & AA\\
\hline
  ReLU & 76.3 & 41.5 & 82.0 & 51.3\\
  PReLU & 78.8 & 40.3  & 82.2 & 51.3\\
    \hline

  ELU & 77.6 & 40.5 &  79.9 & 48.2 \\
  PELU  & 77.4 & 41.1  & 81.6 & 51.2\\
  \hline
  Softplus & 71.9 & 40.2  & 75.5  & 43.8 \\
  PSoftplus & \textbf{81.0} & \textcolor{violet}{41.9} & 82.7 & \textcolor{violet}{53.0} \\
  \hline
  PPRELU & 78.5 & \textcolor{violet}{41.6} & 82.1 & \textcolor{violet}{51.6}\\
  ReBLU & 77.9 & \textcolor{violet}{41.7}  & 82.3 & \textcolor{violet}{52.4} \\
  \hline
  SiLU & 80.5 & \textcolor{violet}{\textbf{43.1}}&82.3 & 51.0 \\
  PSiLU & 77.7 & \textcolor{violet}{42.5}  & \textbf{83.0} & \textcolor{violet}{\textbf{53.2}}\\
  PSSiLU & 77.7 & \textcolor{violet}{41.9}& 82.8 & \textcolor{violet}{52.9}\\
\hline

\end{tabular}
\end{center}
    \caption{Natural and robust accuracy of adversarially trained VGG-16 models of various activation functions with respect to $\ell_{\infty}$ attacks with radius 0.031 generated through AutoAttack. We highlight robust accuracies higher than ReLU in purple.}
    \label{tab:vgg_adv_summary}
\end{table}

\subsection{Results for L2 adversary}
We report results for ResNet-18 models on CIFAR-10 with an L2 adversary in Table \ref{tab:wrn_l2_summary}. We find that the PSiLU achieves the highest robust accuracy both with and without extra data, and PSSiLU has performance comparable to that of PSiLU in both instances.  Additionally, we observe that smooth PAFs (PSoftplus, PSiLU, and PSSiLU) all improve robust accuracy over ReLU even without additional data.
\begin{table}[ht]
    \centering
    \begin{tabular}{|c|c|c|c|c|}
    \hline
    & \multicolumn{2}{|c|}{\textbf{CIFAR-10}} & \multicolumn{2}{|c|}{\textbf{+DDPM-6M}}\\
\hline
Activation & Natural & AA & Natural & AA\\
\hline
  ReLU & 89.6 & 65.1 & 89.4 & 74.4 \\
  PReLU & 84.6 & 57.5 & 88.7 & 71.7\\
    \hline

  ELU & 88.6 & \textcolor{violet}{65.6} & 88.0 &  71.2 \\
  PELU  & 88.7 & 63.8 & 89.1 & 73.9 \\
  \hline
  Softplus & 87.3 & 64.2 & 86.5 & 67.7  \\
  PSoftplus & \textbf{89.8} & \textcolor{violet}{67.1} & 89.7 & \textcolor{violet}{75.3} \\
  \hline
  PPRELU & 88.5 & 63.4 & 88.7 & 73.9\\
  ReBLU & 88.3 & \textcolor{violet}{65.9} & 89.2 & 73.2 \\
  \hline
  SiLU & 87.6 &  64.6 & 89.4 & 73.4\\
  PSiLU & 89.3 & \textcolor{violet}{\textbf{67.7}} & 89.6 & \textcolor{violet}{\textbf{75.8}}\\
  PSSiLU & 89.6 & \textcolor{violet}{67.3} & \textbf{89.9} & \textcolor{violet}{75.6} \\
\hline
    \end{tabular}
    \caption{Natural and robust accuracy of adversarially trained WRN-28-10 models on CIFAR-10 with respect to $\ell_2$ attacks with radius 0.5 generated through AutoAttack. We highlight robust accuracies higher than ReLU in purple.}
    \label{tab:wrn_l2_summary}
\end{table}

\subsection{Results on TRADES trained models}
\label{app:trades}
We report results for ResNet-18 models trained with TRADES adversarial training in Table \ref{tab:TRADES_adv_summary}.  We find that PSSiLU is able to obtain the highest accuracy on CIFAR-10 without additional data, outperforming ReLU by 0.9\%.  With additional data, PSiLU obtains the highest robust accuracy (PSSiLU obtains comparable robustness), outperforming ReLU by 1.9\%.  This is consistent with our observation that the PSSiLU family is able to achieve high robustness.  Additionally, we find that PSoftplus, PSiLU, and PSSiLU all improve over ReLU even without additional data during training.  This is consistent with our observation that smooth PAFs often outperform ReLU.
\begin{table}[ht]
\centering
    \begin{center}
\begin{tabular}{ |c|c|c|c|c| } 
\hline
& \multicolumn{2}{|c|}{\textbf{CIFAR-10}} & \multicolumn{2}{|c|}{\textbf{+DDPM-6M}}\\
\hline
Activation & Natural & AA & Natural & AA\\
\hline
  ReLU & 83.1 & 48.5 & 82.2 & 56.2 \\
  PReLU & 79.6 &  45.4 & \textbf{82.8} & 56.1\\
    \hline
  ELU & 77.5 & 44.5 & 77.8 & 50.8  \\
  PELU  & 82.6 & 47.0 & 82.4 & 55.8 \\
  \hline
  Softplus & 73.9 & 41.4 & 76.7 & 47.5  \\
  PSoftplus & 81.6 & \textcolor{violet}{49.1} & 82.6 & \textcolor{violet}{57.2} \\
  \hline
  PPRELU & \textbf{83.5 }& 47.7 & 82.3 & 55.8 \\
  ReBLU & 82.0 & 46.7 & 81.7 &54.9 \\
  \hline
  SiLU & 81.1 & \textcolor{violet}{49.0} & 79.5 &  53.7\\
  PSiLU & 81.7 & \textcolor{violet}{49.3} & \textbf{82.8} &\textbf{\textcolor{violet}{58.1}}\\
  PSSiLU & 81.2 & \textbf{\textcolor{violet}{49.4}} & 82.7 & \textcolor{violet}{57.9} \\
\hline

\end{tabular}
\end{center}
    \caption{Natural and robust accuracy of TRADES adversarially trained ResNet-18 models of various activation functions with respect to $\ell_{\infty}$ attacks with radius 0.031 generated through AutoAttack. We highlight robust accuracies higher than ReLU in purple.}
    \label{tab:TRADES_adv_summary}
\end{table}

\subsection{Results on Additional Datasets}
%\sdcomment{psoftplus, importance of parametric}
%\sdcomment{outperforming relu}
In Table \ref{tab:imagenette}, we present the results for ResNet-18 models trained on ImageNette and WRN-28-10 models trained on CIFAR-100. We find that on ImageNette, SiLU achieves the highest robust perfomrance, which is consistent with our finding that members of the PSSiLU family are able to achieve high robust accuracy.  We find that PSoftplus achieves the highest robust accuracy on CIFAR-100, but both PSiLU and PSSiLU are able to achieve comparable robust accuracy.  Additionally, we find that across both datasets, PSoftplus, PSiLU, and PSSiLU are able to outperform ReLU, further emphasizing our finding that smooth PAFs generally improve over ReLU even without extra data.

We note that we do not use additional data when training for ImageNette and CIFAR-100.  The peformance of PAFs may further improve if additional data for ImageNette and CIFAR-100 is present.

%We find that the results are consistent with our findings that members of the PSSiLU family are able to achieve high robust accuracy.  We find that on ImageNette, SiLU obtains the highest robust accuracy, suggesting that regularization and training with additional data may improve the performance of PSiLU ad PSSiLU. On CIFAR-100, we find that while PSiLU and PSSiLU do not obtain the highest robust accuracy, they still achieve accuracy in the top 3 and improve over ReLU.  The peformance of PSSiLU and PSiLU may improve if additional data for ImageNette and CIFAR-100 is present.

\begin{table}[ht]
\centering
    \begin{center}
\begin{tabular}{ |c|c|c|c|c| } 
\hline
& \multicolumn{2}{|c|}{\textbf{ImageNette}}& \multicolumn{2}{|c|}{\textbf{CIFAR-100}}\\
\hline
Activation & Natural & AA & Natural & AA\\
\hline
  ReLU & 88.6 & 60.5 & 59.6 &23.5 \\
  PReLU & 69.7 & 59.5 & 57.5 & 21.6\\
    \hline

  ELU & 87.5 & \textcolor{violet}{61.2} & 58.1 &\textcolor{violet}{24.1}\\
  PELU  & 87.2 &57.7 & 58.6 &21.9\\
  \hline
  Softplus & 81.5 &54.8 & 57.1 & 23.1\\
  PSoftplus & \textbf{88.9} & \textcolor{violet}{62.8}  & \textbf{60.5}& \textbf{\textcolor{violet}{24.8}} \\
  \hline
  $\pprelu$ &87.4 &59.6 & 58.6 & 22.7\\
  ReBLU & 87.5 &58.3 & 59.5 & \textcolor{violet}{24.3} \\
  \hline
%  SSiLU & 78.9 & 44.0 & 84.2 & 55.7 & 86.7 & 60.2\\
  SiLU & 88.8& \textbf{\textcolor{violet}{64.1}} & 54.3 &22.9 \\
  PSiLU & 88.7 &\textcolor{violet}{62.4} & 59.3 & \textcolor{violet}{24.2}\\
  PSSiLU& 87.5 & \textcolor{violet}{61.2} & 59.7 & \textcolor{violet}{24.2}\\
\hline

\end{tabular}
\end{center}
    \caption{Natural and robust accuracy of adversarially trained models of various activation functions with respect to $\ell_{\infty}$ attacks with radius 0.031 generated through AutoAttack on ImageNette and CIFAR-100. We highlight robust accuracies higher than ReLU in purple.}
    \label{tab:imagenette}
\end{table}

\subsection{Results with TI-500K}
We report results for WRN-28-10 and VGG-16 models trained with TI-500K as a source of additional data in Table \ref{tab:ti500k}.  We find that additional data improves the performance of all activation functions inclucing PAFs.  Additionally, we find that PSiLU improves over ReLU when trained with TI-500K on WRN-28-10 and PSSiLU, PSiLU, and PSoftplus all outperform ReLU when trained with TI-500K on VGG-16.  Additionally, we observe that PSiLU obtains high robust accuracy comparable to the highest accuracy (achieved by ReBLU) on WRN-28-10 with TI-500K while SiLU obtains the highest accuracy on VGG-16 with the addition TI-500K data.  This validates the performance of the PSSiLU family.
\begin{table}[ht]
\centering
    \begin{center}
\begin{tabular}{ |c|c|c|c|c|c|c|c|c|c|c|c|c| } 
\hline
& \multicolumn{4}{|c|}{\textbf{WRN-28-10}} & \multicolumn{4}{|c|}{\textbf{VGG-16}}\\
\hline

& \multicolumn{2}{|c|}{CIFAR-10} & \multicolumn{2}{|c|}{+TI-500K}& \multicolumn{2}{|c|}{CIFAR-10} & \multicolumn{2}{|c|}{+TI-500K} \\
\hline
Activation & Natural & AA & Natural & AA & Natural & AA & Natural & AA \\
\hline
  ReLU & 83.4 & 46.0 & 89.4  &55.5  & 76.3 & 41.5 & 82.9 & 46.7 \\
  PReLU & 82.8  &43.6  & 89.5  &54.2 & 78.8 & 40.3  &  83.4 & \textcolor{violet}{47.0}\\
    \hline

  ELU & 79.7  &45.9  & 83.7  &50.7 & 77.6 & 40.5  & 82.3 &  45.7 \\
  PELU  & 83.3 & 43.9 & \textbf{90.3} & 54.4 & 77.4 & 41.1 & 83.3 & 46.6\\
  \hline
  Softplus & 80.0  & 44.4 & 80.0  & 45.2 & 71.9 & 40.2 & 77.3 & 43.3\\
  PSoftplus  & 82.9 & \textcolor{violet}{46.7} & 88.9 & 55.1& \textbf{81.0} & \textcolor{violet}{41.9} & \textbf{85.7} & \textcolor{violet}{48.7} \\
  \hline
  $\pprelu$ &81.7& 45.1  & 88.9 & 54.7 & 78.5 & \textcolor{violet}{41.6}& 85.1 & 46.6  \\
  ReBLU  &83.2 & \textcolor{violet}{46.9} & 89.5  & \textbf{\textcolor{violet}{56.2}} & 77.9 & \textcolor{violet}{41.7} & 82.9 & \textcolor{violet}{48.0}\\
  \hline
  SiLU  & 84.2  & \textcolor{violet}{47.5} & 87.9 & 54.8  & 80.5 & \textcolor{violet}{\textbf{43.1}}& 85.3 & \textbf{\textcolor{violet}{48.9}} \\
  PSiLU & 82.4 &\textcolor{violet}{47.0} & 89.9 & \textcolor{violet}{56.1} & 77.7 & \textcolor{violet}{42.5}  & 84.7 & \textcolor{violet}{48.1}\\
  PSSiLU & \textbf{86.0} & \textcolor{violet}{\textbf{48.3}}  & 86.4  &  53.8 & 77.7 & \textcolor{violet}{41.9} & 83.9 & \textcolor{violet}{47.8} \\
\hline

\end{tabular}
\end{center}
    \caption{Natural and robust accuracy of PGD adversarially trained models trained on CIFAR-10 and CIFAR-10+TI-500K with respect to $\ell_{\infty}$ attacks with radius 0.031. We highlight robust accuracies higher than ReLU in purple.}
    \label{tab:ti500k}
\end{table}

\subsection{Fixing $\beta$ on PSSiLU}
\label{app:searchbeta}
Unlike other parametric activation functions tested, PSSiLU has 2 learnable parameters.  We experiment with fixing the value of $\beta$ on PSSiLU so that $\alpha$ is the only learnable parameter.  Figure \ref{fig:ssilu_beta} shows the trend for adversarial difficulty over $\beta$ when $\alpha$ is fixed to 1.  We find after about $\beta=0.3$, we do not see much improvement from increasing the value of $\beta$.  We set $\beta$ to 0.3 and trained another set of ResNet-18 models using PSSiLU.  The results are shown in Table \ref{tab:fixed_beta}.

\begin{table}[h]
    \centering
    \begin{tabular}{ |c|c|c|c|c| } 
\hline
& \multicolumn{2}{|c|}{\textbf{CIFAR-10}} &  \multicolumn{2}{|c|}{\textbf{+DDPM-6M}}\\
\hline
Activation & Natural & AA & Natural & AA\\
\hline
    ReLU & \textbf{82.3} & \textbf{44.6} &  82.8 & 53.7\\
    PSSiLU & 79.2 & 42.7 & \textbf{84.5} & \textbf{56.8} \\
\hline
    \end{tabular}
    \caption{Natural and robust accuracy of PSSiLU model with $\beta$ fixed at 0.3 with respect to L-infinity attacks with radius 0.031 generated by AutoAttack.}
    \label{tab:fixed_beta}
\end{table}
We find that even with fixed $\beta$, we are able to achieve high robust accuracy on CIFAR-10 when combined with DDPM-6M; however, it is not as high as with $\beta$ as a learnable parameter.

\subsection{Impact of regularization on PSSiLU performance}
\label{app:pssilu_reg}
We vary the strength of regularization ($\lambda$) on $\beta$.  We observe that there is a significant jump in robust performance from no regularization $\lambda = 0$ to $\lambda = 0.1$ suggesting that PSSiLU needs regularization to perform well.  This makes sense because in our formulation for PSSiLU, we have the constraint that $\beta < 1$ which allows PSSiLU to maintain a ReLU-like shape. We find that the best performing model is produced when $\lambda=10$.
\begin{figure}[h]
    \centering
    \includegraphics[width=0.5\textwidth]{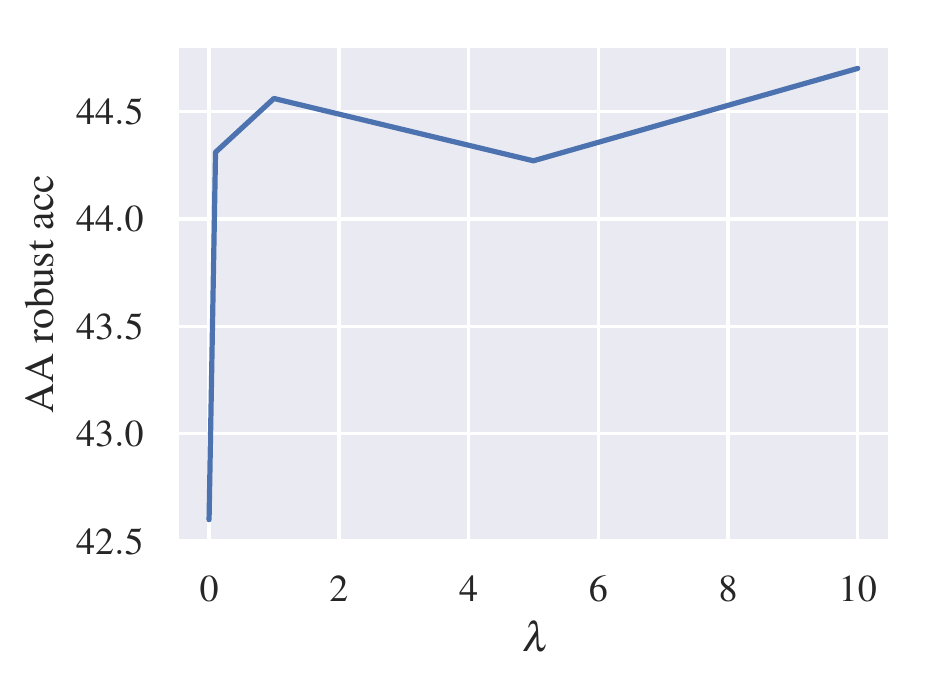}
    \caption{Impact of regularization strength $\lambda$ on $\beta$ parameter on AutoAttack robust accuracy of PGD adversarially trained ResNet-18 model.}
    \label{fig:beta_reg}
\end{figure}

\subsection{Learned shapes for $\pprelu$ and ReBLU}
\begin{figure}[h]
    \centering
    \includegraphics[width=0.75\textwidth]{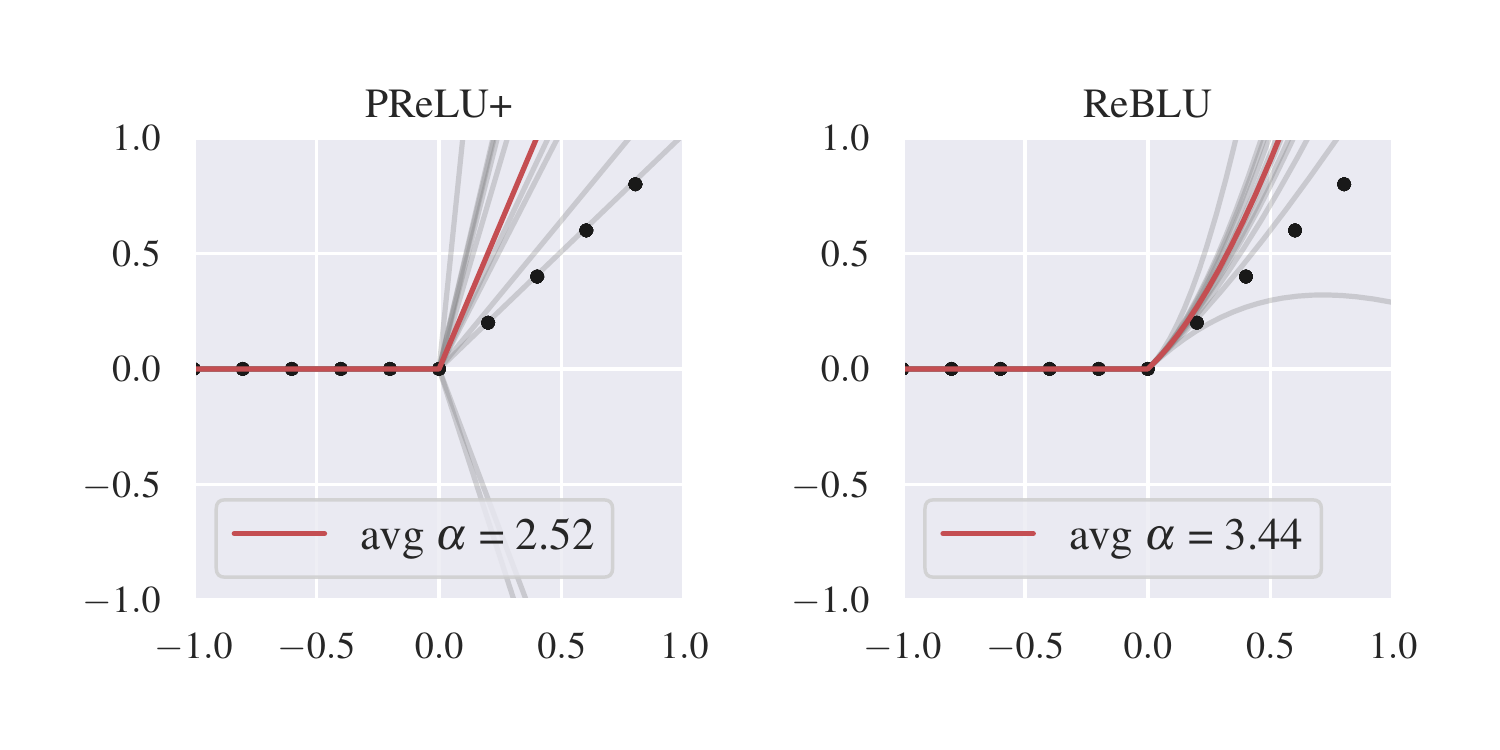}
    \caption{Learned shapes of $\pprelu$ and ReBLU activation functions across all 11 models trained using PGD adversarial training.  Each gray line represents the shape learned by a single model.  The red line represents the average of the learned $\alpha$s across all models.}
    \label{fig:pprelu_pblu_learned_shape}
\end{figure}

We present the learned shapes of $\pprelu$ and ReBLU in Figure \ref{fig:pprelu_pblu_learned_shape}.  We find that these activation functions generally optimize so that the slope in the positive region is positive.  However, we find that this trend is not consistent across dataset and architecture.  For instance in Figure \ref{fig:pprelu_pblu_learned_shape}, we can see several models which optimize towards negative values of $\alpha$, leading to negative slope on positive inputs on $\pprelu$ and a downward curve on ReBLU.


\nocite{*}
\begin{thebibliography}{49}
\providecommand{\natexlab}[1]{#1}
\providecommand{\url}[1]{\texttt{#1}}
\expandafter\ifx\csname urlstyle\endcsname\relax
  \providecommand{\doi}[1]{doi: #1}\else
  \providecommand{\doi}{doi: \begingroup \urlstyle{rm}\Url}\fi

\bibitem[Andriushchenko et~al.(2020)Andriushchenko, Croce, Flammarion, and
  Hein]{andriushchenko2020square}
Maksym Andriushchenko, Francesco Croce, Nicolas Flammarion, and Matthias Hein.
\newblock Square attack: a query-efficient black-box adversarial attack via
  random search.
\newblock In \emph{European Conference on Computer Vision}, pp.\  484--501.
  Springer, 2020.

\bibitem[Athalye et~al.(2018)Athalye, Carlini, and
  Wagner]{athalye2018obfuscated}
Anish Athalye, Nicholas Carlini, and David Wagner.
\newblock Obfuscated gradients give a false sense of security: Circumventing
  defenses to adversarial examples.
\newblock In \emph{International conference on machine learning}, pp.\
  274--283. PMLR, 2018.

\bibitem[Brendel et~al.(2018)Brendel, Rauber, and Bethge]{brendel2017decision}
Wieland Brendel, Jonas Rauber, and Matthias Bethge.
\newblock Decision-based adversarial attacks: Reliable attacks against
  black-box machine learning models.
\newblock In \emph{International Conference on Learning Representations}, 2018.

\bibitem[Bubeck \& Sellke(2021)Bubeck and Sellke]{bubeck2021universal}
S{\'e}bastien Bubeck and Mark Sellke.
\newblock A universal law of robustness via isoperimetry.
\newblock \emph{arXiv preprint arXiv:2105.12806}, 2021.

\bibitem[Carlini \& Wagner(2017)Carlini and Wagner]{carlini2017towards}
Nicholas Carlini and David Wagner.
\newblock Towards evaluating the robustness of neural networks.
\newblock In \emph{2017 IEEE symposium on security and privacy (sp)}, pp.\
  39--57. IEEE, 2017.

\bibitem[Carmon et~al.(2019)Carmon, Raghunathan, Schmidt, Duchi, and
  Liang]{carmon2019unlabeled}
Yair Carmon, Aditi Raghunathan, Ludwig Schmidt, John~C Duchi, and Percy~S
  Liang.
\newblock Unlabeled data improves adversarial robustness.
\newblock In \emph{Advances in Neural Information Processing Systems}, pp.\
  11192--11203, 2019.

\bibitem[Clevert et~al.(2016)Clevert, Unterthiner, and
  Hochreiter]{clevert2015fast}
Djork{-}Arn{\'{e}} Clevert, Thomas Unterthiner, and Sepp Hochreiter.
\newblock Fast and accurate deep network learning by exponential linear units
  (elus).
\newblock In Yoshua Bengio and Yann LeCun (eds.), \emph{4th International
  Conference on Learning Representations, {ICLR} 2016, San Juan, Puerto Rico,
  May 2-4, 2016, Conference Track Proceedings}, 2016.
\newblock URL \url{http://arxiv.org/abs/1511.07289}.

\bibitem[Croce \& Hein(2020)Croce and Hein]{croce2020reliable}
Francesco Croce and Matthias Hein.
\newblock Reliable evaluation of adversarial robustness with an ensemble of
  diverse parameter-free attacks.
\newblock In \emph{International conference on machine learning}, pp.\
  2206--2216. PMLR, 2020.

\bibitem[Croce et~al.(2020)Croce, Andriushchenko, Sehwag, Debenedetti,
  Flammarion, Chiang, Mittal, and Hein]{croce2020robustbench}
Francesco Croce, Maksym Andriushchenko, Vikash Sehwag, Edoardo Debenedetti,
  Nicolas Flammarion, Mung Chiang, Prateek Mittal, and Matthias Hein.
\newblock Robustbench: a standardized adversarial robustness benchmark.
\newblock \emph{arXiv preprint arXiv:2010.09670}, 2020.

\bibitem[Dugas et~al.(2001)Dugas, Bengio, B{\'e}lisle, Nadeau, and
  Garcia]{dugas2001incorporating}
Charles Dugas, Yoshua Bengio, Fran{\c{c}}ois B{\'e}lisle, Claude Nadeau, and
  Ren{\'e} Garcia.
\newblock Incorporating second-order functional knowledge for better option
  pricing.
\newblock \emph{Advances in neural information processing systems}, pp.\
  472--478, 2001.

\bibitem[Glorot et~al.(2011)Glorot, Bordes, and Bengio]{glorot2011deep}
Xavier Glorot, Antoine Bordes, and Yoshua Bengio.
\newblock Deep sparse rectifier neural networks.
\newblock In \emph{Proceedings of the fourteenth international conference on
  artificial intelligence and statistics}, pp.\  315--323. JMLR Workshop and
  Conference Proceedings, 2011.

\bibitem[Godfrey(2019)]{8913972}
Luke~B. Godfrey.
\newblock An evaluation of parametric activation functions for deep learning.
\newblock In \emph{2019 IEEE International Conference on Systems, Man and
  Cybernetics (SMC)}, pp.\  3006--3011, 2019.
\newblock \doi{10.1109/SMC.2019.8913972}.

\bibitem[Goodfellow et~al.(2015)Goodfellow, Shlens, and
  Szegedy]{goodfellow2014explaining}
Ian~J. Goodfellow, Jonathon Shlens, and Christian Szegedy.
\newblock Explaining and harnessing adversarial examples.
\newblock In Yoshua Bengio and Yann LeCun (eds.), \emph{3rd International
  Conference on Learning Representations, {ICLR} 2015, San Diego, CA, USA, May
  7-9, 2015, Conference Track Proceedings}, 2015.
\newblock URL \url{http://arxiv.org/abs/1412.6572}.

\bibitem[Gowal et~al.(2020)Gowal, Qin, Uesato, Mann, and
  Kohli]{gowal2020uncovering}
Sven Gowal, Chongli Qin, Jonathan Uesato, Timothy Mann, and Pushmeet Kohli.
\newblock Uncovering the limits of adversarial training against norm-bounded
  adversarial examples.
\newblock \emph{arXiv preprint arXiv:2010.03593}, 2020.

\bibitem[He et~al.(2015)He, Zhang, Ren, and Sun]{he2015delving}
Kaiming He, Xiangyu Zhang, Shaoqing Ren, and Jian Sun.
\newblock Delving deep into rectifiers: Surpassing human-level performance on
  imagenet classification.
\newblock In \emph{Proceedings of the IEEE international conference on computer
  vision}, pp.\  1026--1034, 2015.

\bibitem[He et~al.(2016)He, Zhang, Ren, and Sun]{he2016deep}
Kaiming He, Xiangyu Zhang, Shaoqing Ren, and Jian Sun.
\newblock Deep residual learning for image recognition.
\newblock In \emph{Proceedings of the IEEE Conference on Computer Vision and
  Pattern Recognition}, pp.\  770--778, 2016.

\bibitem[Ho et~al.(2020)Ho, Jain, and Abbeel]{ho2020denoising}
Jonathan Ho, Ajay Jain, and Pieter Abbeel.
\newblock Denoising diffusion probabilistic models.
\newblock In Hugo Larochelle, Marc'Aurelio Ranzato, Raia Hadsell,
  Maria{-}Florina Balcan, and Hsuan{-}Tien Lin (eds.), \emph{Advances in Neural
  Information Processing Systems 33: Annual Conference on Neural Information
  Processing Systems 2020, NeurIPS 2020, December 6-12, 2020, virtual}, 2020.
\newblock URL
  \url{https://proceedings.neurips.cc/paper/2020/hash/4c5bcfec8584af0d967f1ab10179ca4b-Abstract.html}.

\bibitem[Howard()]{howardimagenette}
Jeremy Howard.
\newblock Imagewang.
\newblock URL \url{https://github.com/fastai/imagenette/}.

\bibitem[Jordan \& Dimakis(2020)Jordan and Dimakis]{jordan2020exactly}
Matt Jordan and Alexandros~G Dimakis.
\newblock Exactly computing the local lipschitz constant of relu networks.
\newblock \emph{arXiv preprint arXiv:2003.01219}, 2020.

\bibitem[Krizhevsky et~al.(2009)Krizhevsky, Hinton,
  et~al.]{krizhevsky2009learning}
Alex Krizhevsky, Geoffrey Hinton, et~al.
\newblock Learning multiple layers of features from tiny images.
\newblock 2009.

\bibitem[Loshchilov \& Hutter(2017)Loshchilov and Hutter]{loshchilov2016sgdr}
Ilya Loshchilov and Frank Hutter.
\newblock {SGDR:} stochastic gradient descent with warm restarts.
\newblock In \emph{5th International Conference on Learning Representations,
  {ICLR} 2017, Toulon, France, April 24-26, 2017, Conference Track
  Proceedings}, 2017.
\newblock URL \url{https://openreview.net/forum?id=Skq89Scxx}.

\bibitem[Madry et~al.(2018)Madry, Makelov, Schmidt, Tsipras, and
  Vladu]{madry2017towards}
Aleksander Madry, Aleksandar Makelov, Ludwig Schmidt, Dimitris Tsipras, and
  Adrian Vladu.
\newblock Towards deep learning models resistant to adversarial attacks.
\newblock In \emph{International Conference on Learning Representations}, 2018.

\bibitem[Moosavi-Dezfooli et~al.(2019)Moosavi-Dezfooli, Fawzi, Uesato, and
  Frossard]{moosavi2019robustness}
Seyed-Mohsen Moosavi-Dezfooli, Alhussein Fawzi, Jonathan Uesato, and Pascal
  Frossard.
\newblock Robustness via curvature regularization, and vice versa.
\newblock In \emph{Proceedings of the IEEE/CVF Conference on Computer Vision
  and Pattern Recognition}, pp.\  9078--9086, 2019.

\bibitem[Pang et~al.(2020)Pang, Yang, Dong, Su, and Zhu]{pang2020bag}
Tianyu Pang, Xiao Yang, Yinpeng Dong, Hang Su, and Jun Zhu.
\newblock Bag of tricks for adversarial training.
\newblock In \emph{International Conference on Learning Representations}, 2020.

\bibitem[Papernot et~al.(2016)Papernot, McDaniel, and
  Goodfellow]{papernot2016transferability}
Nicolas Papernot, Patrick McDaniel, and Ian Goodfellow.
\newblock Transferability in machine learning: from phenomena to black-box
  attacks using adversarial samples.
\newblock \emph{arXiv preprint arXiv:1605.07277}, 2016.

\bibitem[Pauli et~al.(2021)Pauli, Koch, Berberich, Kohler, and
  Allgower]{pauli2021training}
Patricia Pauli, Anne Koch, Julian Berberich, Paul Kohler, and Frank Allgower.
\newblock Training robust neural networks using lipschitz bounds.
\newblock \emph{IEEE Control Systems Letters}, 2021.

\bibitem[Qin et~al.(2019)Qin, Martens, Gowal, Krishnan, Dvijotham, Fawzi, De,
  Stanforth, and Kohli]{NEURIPS2019_0defd533}
Chongli Qin, James Martens, Sven Gowal, Dilip Krishnan, Krishnamurthy
  Dvijotham, Alhussein Fawzi, Soham De, Robert Stanforth, and Pushmeet Kohli.
\newblock Adversarial robustness through local linearization.
\newblock In H.~Wallach, H.~Larochelle, A.~Beygelzimer, F.~d\textquotesingle
  Alch\'{e}-Buc, E.~Fox, and R.~Garnett (eds.), \emph{Advances in Neural
  Information Processing Systems}, volume~32. Curran Associates, Inc., 2019.
\newblock URL
  \url{https://proceedings.neurips.cc/paper/2019/file/0defd533d51ed0a10c5c9dbf93ee78a5-Paper.pdf}.

\bibitem[Rakin et~al.(2018)Rakin, Yi, Gong, and Fan]{rakin2018defend}
Adnan~Siraj Rakin, Jinfeng Yi, Boqing Gong, and Deliang Fan.
\newblock Defend deep neural networks against adversarial examples via fixed
  and dynamic quantized activation functions.
\newblock \emph{arXiv preprint arXiv:1807.06714}, 2018.

\bibitem[Ramachandran et~al.(2018)Ramachandran, Zoph, and
  Le]{ramachandran2017searching}
Prajit Ramachandran, Barret Zoph, and Quoc~V. Le.
\newblock Searching for activation functions.
\newblock In \emph{6th International Conference on Learning Representations,
  {ICLR} 2018, Vancouver, BC, Canada, April 30 - May 3, 2018, Workshop Track
  Proceedings}, 2018.
\newblock URL \url{https://openreview.net/forum?id=Hkuq2EkPf}.

\bibitem[Rebuffi et~al.(2021)Rebuffi, Gowal, Calian, Stimberg, Wiles, and
  Mann]{rebuffi2021fixing}
Sylvestre{-}Alvise Rebuffi, Sven Gowal, Dan~A. Calian, Florian Stimberg, Olivia
  Wiles, and Timothy~A. Mann.
\newblock Fixing data augmentation to improve adversarial robustness.
\newblock \emph{CoRR}, abs/2103.01946, 2021.
\newblock URL \url{https://arxiv.org/abs/2103.01946}.

\bibitem[Rice et~al.(2020)Rice, Wong, and Kolter]{rice2020overfitting}
Leslie Rice, Eric Wong, and Zico Kolter.
\newblock Overfitting in adversarially robust deep learning.
\newblock In \emph{International Conference on Machine Learning}, pp.\
  8093--8104. PMLR, 2020.

\bibitem[Sehwag et~al.(2021)Sehwag, Mahloujifar, Handina, Dai, Xiang, Chiang,
  and Mittal]{sehwag2021improving}
Vikash Sehwag, Saeed Mahloujifar, Tinashe Handina, Sihui Dai, Chong Xiang, Mung
  Chiang, and Prateek Mittal.
\newblock Improving adversarial robustness using proxy distributions.
\newblock \emph{arXiv preprint arXiv:2104.09425}, 2021.

\bibitem[Simonyan \& Zisserman(2015)Simonyan and Zisserman]{simonyan2015vgg}
Karen Simonyan and Andrew Zisserman.
\newblock Very deep convolutional networks for large-scale image recognition.
\newblock In Yoshua Bengio and Yann LeCun (eds.), \emph{3rd International
  Conference on Learning Representations, {ICLR} 2015, San Diego, CA, USA, May
  7-9, 2015, Conference Track Proceedings}, 2015.
\newblock URL \url{http://arxiv.org/abs/1409.1556}.

\bibitem[Singla et~al.(2021)Singla, Singla, Jacobs, and Feizi]{singla2021low}
Vasu Singla, Sahil Singla, David Jacobs, and Soheil Feizi.
\newblock Low curvature activations reduce overfitting in adversarial training.
\newblock \emph{arXiv preprint arXiv:2102.07861}, 2021.

\bibitem[Szegedy et~al.(2014)Szegedy, Zaremba, Sutskever, Bruna, Erhan,
  Goodfellow, and Fergus]{szegedy2013intriguing}
Christian Szegedy, Wojciech Zaremba, Ilya Sutskever, Joan Bruna, Dumitru Erhan,
  Ian~J. Goodfellow, and Rob Fergus.
\newblock Intriguing properties of neural networks.
\newblock In Yoshua Bengio and Yann LeCun (eds.), \emph{2nd International
  Conference on Learning Representations, {ICLR} 2014, Banff, AB, Canada, April
  14-16, 2014, Conference Track Proceedings}, 2014.
\newblock URL \url{http://arxiv.org/abs/1312.6199}.

\bibitem[Tavakoli et~al.(2020)Tavakoli, Agostinelli, and
  Baldi]{tavakoli2020splash}
Mohammadamin Tavakoli, Forest Agostinelli, and Pierre Baldi.
\newblock Splash: Learnable activation functions for improving accuracy and
  adversarial robustness.
\newblock \emph{arXiv preprint arXiv:2006.08947}, 2020.

\bibitem[Wang et~al.(2018)Wang, Lin, Shi, Zhu, Yin, Bertozzi, and
  Osher]{wang2018adversarial}
Bao Wang, Alex~T Lin, Zuoqiang Shi, Wei Zhu, Penghang Yin, Andrea~L Bertozzi,
  and Stanley~J Osher.
\newblock Adversarial defense via data dependent activation function and total
  variation minimization.
\newblock \emph{arXiv preprint arXiv:1809.08516}, 2018.

\bibitem[Wang et~al.(2019)Wang, Ma, Bailey, Yi, Zhou, and Gu]{wang2019dynamic}
Yisen Wang, Xingjun Ma, James Bailey, Jinfeng Yi, Bowen Zhou, and Quanquan Gu.
\newblock On the convergence and robustness of adversarial training.
\newblock In \emph{International Conference on Machine Learning}, 2019.

\bibitem[Wu et~al.(2020{\natexlab{a}})Wu, Chen, Cai, He, and Gu]{wu2020does}
Boxi Wu, Jinghui Chen, Deng Cai, Xiaofei He, and Quanquan Gu.
\newblock Does network width really help adversarial robustness?
\newblock \emph{arXiv preprint arXiv:2010.01279}, 2020{\natexlab{a}}.

\bibitem[Wu et~al.(2020{\natexlab{b}})Wu, Xia, and Wang]{wu2020adversarial}
Dongxian Wu, Shu-Tao Xia, and Yisen Wang.
\newblock Adversarial weight perturbation helps robust generalization.
\newblock In \emph{NeurIPS}, 2020{\natexlab{b}}.

\bibitem[Xie \& Yuille(2019)Xie and Yuille]{xie2019intriguing}
Cihang Xie and Alan Yuille.
\newblock Intriguing properties of adversarial training at scale.
\newblock In \emph{International Conference on Learning Representations}, 2019.

\bibitem[Xie et~al.(2020)Xie, Tan, Gong, Yuille, and Le]{xie2020smooth}
Cihang Xie, Mingxing Tan, Boqing Gong, Alan Yuille, and Quoc~V Le.
\newblock Smooth adversarial training.
\newblock \emph{arXiv preprint arXiv:2006.14536}, 2020.

\bibitem[Yang et~al.(2020)Yang, Rashtchian, Zhang, Salakhutdinov, and
  Chaudhuri]{yang2020closer}
Yao-Yuan Yang, Cyrus Rashtchian, Hongyang Zhang, Russ~R Salakhutdinov, and
  Kamalika Chaudhuri.
\newblock A closer look at accuracy vs. robustness.
\newblock In \emph{NeurIPS}, 2020.

\bibitem[Zagoruyko \& Komodakis(2016)Zagoruyko and
  Komodakis]{zagoruyko2016wide}
Sergey Zagoruyko and Nikos Komodakis.
\newblock Wide residual networks.
\newblock \emph{arXiv preprint arXiv:1605.07146}, 2016.

\bibitem[Zantedeschi et~al.(2017)Zantedeschi, Nicolae, and
  Rawat]{zantedeschi2017efficient}
Valentina Zantedeschi, Maria-Irina Nicolae, and Ambrish Rawat.
\newblock Efficient defenses against adversarial attacks.
\newblock In \emph{Proceedings of the 10th ACM Workshop on Artificial
  Intelligence and Security}, pp.\  39--49, 2017.

\bibitem[Zhang et~al.(2019)Zhang, Yu, Jiao, Xing, Ghaoui, and
  Jordan]{zhang2019theoretically}
Hongyang Zhang, Yaodong Yu, Jiantao Jiao, Eric~P. Xing, Laurent~El Ghaoui, and
  Michael~I. Jordan.
\newblock Theoretically principled trade-off between robustness and accuracy.
\newblock In Kamalika Chaudhuri and Ruslan Salakhutdinov (eds.),
  \emph{Proceedings of the 36th International Conference on Machine Learning,
  {ICML} 2019, 9-15 June 2019, Long Beach, California, {USA}}, volume~97 of
  \emph{Proceedings of Machine Learning Research}, pp.\  7472--7482. {PMLR},
  2019.
\newblock URL \url{http://proceedings.mlr.press/v97/zhang19p.html}.

\bibitem[Zhang et~al.(2020{\natexlab{a}})Zhang, Xu, Han, Niu, Cui, Sugiyama,
  and Kankanhalli]{zhang2020fat}
Jingfeng Zhang, Xilie Xu, Bo~Han, Gang Niu, Lizhen Cui, Masashi Sugiyama, and
  Mohan Kankanhalli.
\newblock Attacks which do not kill training make adversarial learning
  stronger.
\newblock In \emph{ICML}, 2020{\natexlab{a}}.

\bibitem[Zhang et~al.(2020{\natexlab{b}})Zhang, Zhu, Niu, Han, Sugiyama, and
  Kankanhalli]{zhang2020geometry}
Jingfeng Zhang, Jianing Zhu, Gang Niu, Bo~Han, Masashi Sugiyama, and Mohan
  Kankanhalli.
\newblock Geometry-aware instance-reweighted adversarial training.
\newblock In \emph{International Conference on Learning Representations},
  2020{\natexlab{b}}.

\bibitem[Zhao \& Griffin(2016)Zhao and Griffin]{zhao2016suppressing}
Qiyang Zhao and Lewis~D. Griffin.
\newblock Suppressing the unusual: towards robust cnns using symmetric
  activation functions.
\newblock \emph{CoRR}, abs/1603.05145, 2016.
\newblock URL \url{http://arxiv.org/abs/1603.05145}.

\end{thebibliography}
\end{document}